\documentclass[10pt,twocolumn,letterpaper]{article}

\usepackage{cvpr}              % To produce the CAMERA-READY version
\usepackage[accsupp]{axessibility}
\usepackage{microtype}
\usepackage{graphicx}
\usepackage{multirow}
\usepackage{booktabs}
\usepackage{colortbl}
\usepackage{subfloat}
\usepackage{floatrow}
\usepackage[small]{caption}
\usepackage{float}

\newfloatcommand{capbtabbox}{table}[][\FBwidth]
\newfloatcommand{capbfigbox}{figure}[][\FBwidth]

\usepackage{algorithm}
\usepackage{algorithmicx}
\usepackage[noend]{algpseudocode}

\usepackage{pifont}
\usepackage{bbding}
\usepackage{fontawesome}
\usepackage[dvipsnames]{xcolor}

\definecolor{cvprblue}{rgb}{0.21,0.49,0.74}
\definecolor{mydarkred}{RGB}{139, 0, 0}
\definecolor{blank}{HTML}{E6E6E6}
\usepackage[pagebackref,breaklinks,colorlinks,citecolor=cvprblue]{hyperref}

\newcommand{\cmark}{\color{blue}{\ding{51}}}
\newcommand{\xmark}{\color{red}{\ding{55}}}

\newcommand\lft{\mathopen{}\left}
\newcommand\rgt{\aftergroup\mathclose\aftergroup{\aftergroup}\right}

\def\paperID{15226} % *** Enter the Paper ID here
\def\confName{CVPR}
\def\confYear{2024}

\title{Embodied Multi-Modal Agent trained by an LLM from a Parallel TextWorld}

\author{
Yijun Yang$^{1,5,4}$
\quad Tianyi Zhou$^{2}$
\quad Kanxue Li$^{3}$
\quad Dapeng Tao$^{3}$
\quad Lusong Li$^{4}$ \\
\quad Li Shen$^{4,*}$
\quad Xiaodong He$^{4}$
\quad Jing Jiang$^{5}$
\quad Yuhui Shi$^{1,}$\thanks{Corresponding author} \\[0.5mm]
$^{1}$Southern University of Science and Technology \quad $^{2}$University of Maryland, College Park \\ 
$^{3}$Yunnan University \quad $^{4}$JD Explore Academy \quad $^{5}$University of Technology Sydney\\
{\tt\small \url{https://github.com/stevenyangyj/Emma-Alfworld}}
}

\begin{document}
\maketitle

\begin{abstract}
While large language models (LLMs) excel in a simulated world of texts, they struggle to interact with the more realistic world without perceptions of other modalities such as visual or audio signals. Although vision-language models (VLMs) integrate LLM modules (1) aligned with static image features, and (2) may possess prior knowledge of world dynamics (as demonstrated in the text world), they have not been trained in an embodied visual world and thus cannot align with its dynamics. On the other hand, training an embodied agent in a noisy visual world without expert guidance is often challenging and inefficient. 
In this paper, we train a VLM agent living in a visual world using an LLM agent excelling in a parallel text world. 
% \ls{this is motivation. Move it to 2-th sentence and show the drawbacks and challenges of the existing methods.}. \ls{We directly show how and why our method works.} 
Specifically, we distill LLM's reflection outcomes (improved actions by analyzing mistakes) in a text world's tasks to finetune the VLM on the same tasks of the visual world, resulting in an \textbf{E}mbodied \textbf{M}ulti-\textbf{M}odal \textbf{A}gent (\textbf{EMMA}) quickly adapting to the visual world dynamics. Such cross-modality imitation learning between the two parallel worlds is achieved by a novel DAgger-DPO algorithm, enabling EMMA to generalize to a broad scope of new tasks without any further guidance from the LLM expert. Extensive evaluations on the ALFWorld benchmark's diverse tasks highlight EMMA's superior performance to SOTA VLM-based agents, e.g., 20\%-70\% improvement in the success rate.%\looseness-1
% While large language models (LLMs) have been increasingly used to control an embodied agent to interact with external environments, the seamless integration of embodied perceptual modalities, such as visual data, within LLMs remains an open problem. The emergence of vision-language models (VLMs), with their inherent capability to process both visual and linguistic data, offers a potential solution. However, the drastic variability of data distribution across different environments and tasks poses challenges, resulting in the suboptimal performance of VLM-based agents fine-tuned on a fixed demonstration dataset. We address it by \textbf{E}mbodied \textbf{M}ulti-\textbf{M}odal \textbf{A}gent (\textbf{EMMA}), which aims at adapting the VLM-based  agent to any dynamic environment efficiently. The core idea is to train the agent by imitating a reflective LLM expert who can verbally reflect on the agent's mistakes and then leverage the reflective text to prompt itself to generate better actions that the agent will imitate in subsequent trials. Once trained, the VLM agent can complete a broad scope of tasks without further any guidance from the LLM expert. Our evaluations on the ALFWorld benchmark highlight EMMA's superior performance compared to state-of-the-art VLM-based agents across diverse tasks.
\end{abstract}

\vspace{-1em}
\section{Introduction}
\label{sec:intro}

% \ty{Recent advances of LLMs and VLMs enable us to create more powerful agents mainly due to the reasoning capability and prior knowledge of language models.} 

Embodied multi-modal agents have been acknowledged as a pivotal stride towards achieving Artificial General Intelligence (AGI), as they encompass the potential for a broad scope of intelligent activities~\citep{wooldridge1995intelligent,hutter2004universal}. The rising of foundation models brings a glimmer of hope for constructing such agents~\citep{corr/chunyuan_vlm_survey,icml/blip2,corr/instructblip,corr/minigpt4,corr/llava,yang2023continual,liang2024module}, and notable efforts from the community have tried to harness them in many decision-making scenarios, e.g., autonomous driving~\citep{corr/drive_like_human,xu2023drivegpt4}, daily household robots~\citep{icml/palm-e,rss/rt-1,corr/rt-2}, and complex manipulation tasks~\citep{huang2023voxposer,corr/embodiedgpt,icml/distill_vlm,arxiv/VIMA}.

% \ty{The challenges of LLMs/VLMs when being used as agents.} 
LLMs can perform as reflex agents interacting with a text world via verbalized descriptions of the world states and textual actions. In addition, they exhibit great potential in planning, reflection, and reward shaping. This is mainly due to their prior knowledge and semantic abstraction of the world. However, LLM-based agents cannot be directly applied in a visual world. 
% On the other hand, it is still challenging to train an agent in a visual world due to its complexity, the lack of semantic structures in the perceived pixels, and the noises of the visual signals, which cannot be fully handled by the current vision models such as ViT~\citep{dosovitskiy2020vit}. 
Although current vision-language models (VLMs) try to align LLMs with the visual modality, their pretraining only focuses on static alignment between image-text pairs so the resulting agents have not been aligned well with the dynamics of the visual world.
As shown in Fig.~\ref{fig:teaser} (a), even the privileged VLM, e.g., GPT-4V~\citep{gpt4v}, fails to accomplish tasks in an embodied ALFWorld environment~\citep{iclr/alfworld}. In such a zero-shot setting, GPT-4V tends to mainly rely on its language prior of these objects detected in the current step, rather than the alignment between the visual input and the environment dynamics conditioned on the task instruction. 
% Hence, how can we finetune a VLM to be an embodied agent with dynamic alignment to the visual world?
% Furthermore, various VLMs cannot use the most powerful LLMs since they are close-sourced. Can we transfer the capbility of API LLM reflex agents in a text world to a VLM agent in a visual world?

% \ty{The motivations and overarching goal of training a VLM agent in a visual world (dynamic alignment).} 
In this paper, we study how to train a VLM towards an embodied agent in a visual world by aligning it with the visual world dynamics and distilling the skills of an LLM agent in a parallel text world. Our overarching goal is to build such an \textbf{E}mbodied \textbf{M}ulti-\textbf{M}odal \textbf{A}gent (\textbf{EMMA}) that can take a textual task instruction (e.g., from human users) and pixel observations of the state per step to produce a sequence of actions leading to the efficient completion of the task. This is a challenging problem due to (1) the sparsity of task reward~\citep{andrychowicz2017hindsight,fang2019curriculum,ao2021co,ao2022eat}, (2) noisy visual representations, (3) the hallucination of VLMs~\citep{wang2024mementos}, and (4) the misalignment of VLM's static representations to the visual world dynamics. While offline distillation and imitation from a powerful LLM agent in a parallel text world can potentially overcome the former two challenges~\citep{zhong2024policy}, effectively mitigating the remaining challenges necessitates online finetuning of the VLM agent within an interactive and visual world~\citep{corr/offlinerl_survey,iclr/p3,guan2023voce}.

% \ty{Our main technical contributions and novelty.}
To this end, we finetune a VLM agent by imitation learning from an LLM expert (e.g., built on ChatGPT) launched in a parallel text world on the same tasks. 
Specifically, in each step of EMMA interacting with the visual world, 
% EMMA takes a textual description of a task and pixel observation as its input state per step to generate a sequence of actions using the VLM. 
we convert its visual observation into an equivalent textual description sent to the LLM agent, which produces an action for EMMA to imitate. Such cross-modality interactive imitation learning is based on DAgger~\citep{aistats/dagger}, which overcomes the cumulative errors and distribution shifts caused by behavior cloning (BC). As depicted in Fig.~\ref{fig:teaser} (b), an InstructBLIP~\citep{corr/instructblip} agent finetuned by BC on 170K expert demonstrations produced by a rule-based expert in the visual world still fails to take correct actions based on visual observations. We further improve the DAgger's objective to be the direct preference optimization (DPO)~\citep{rafailov2024direct}, which maximizes the preference of LLM-expert's action (positive) over VLM-student's action (negative) in each interaction step. 
To collect better teaching signals retrospective to the VLM student's actions, the LLM expert is composed of an LLM actor prompted to output expert actions, and an LLM critic prompted for reflection feedback on the VLM agent's historical trajectories. We maintain a long-term memory storing the feedback, which is then used to induce the LLM actor to improve actions for imitation in future episodes.\looseness-1
% Moveover, while API LLM agent can provide expert by directly reasoning on the environmental context, these actions may be suboptimal due to sparse environmental feedback or defective in-context instructions~\citep{chen2023instructzero}. 
\begin{figure}[t]
    \centering
    \includegraphics[width=\linewidth]{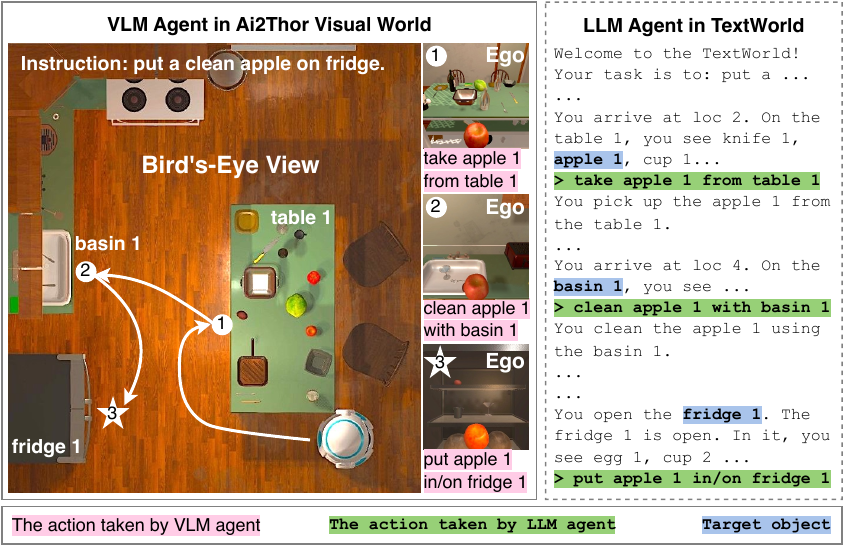}
    % \vspace{-5pt}
    \caption{\textbf{An example of tasks generated for the two parallel worlds.} A VLM agent in the visual world and an LLM agent in the text world as household robots instructed to clean an apple and then put it into the fridge. Zoom in for more details.}
    \label{fig:parallel_worlds}
\end{figure}

% \ty{Our main discoveries and empirical results/achievements.}
We evaluate EMMA and compare it with vision-only agents, LLM agents, and VLM agents deployed to the ALFWorld benchmark~\citep{iclr/alfworld}, which includes numerous tasks in both visual and textual environments. Extensive evaluations highlight that EMMA substantially outperforms state-of-the-art (SOTA) VLM agents in visual-only environments by 20\%-70\% in terms of success rate. In addition, EMMA is the only VLM agent that can generalize to open-vocabulary and free-form test tasks, shedding novel insights on using LLM feedback to train more versatile and generalizable embodied agents in multi-modality environments. \looseness-1

\begin{figure*}[t!]
% \vspace{-0.3em}
    \centering
    \includegraphics[width=0.95\textwidth]{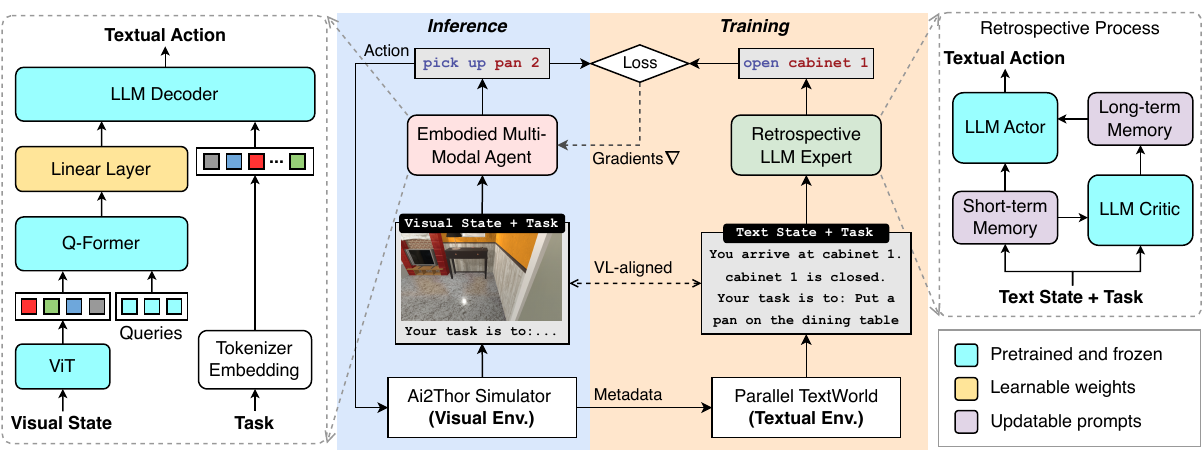}
    % \vspace{-1.5em}
    \caption{\looseness=-1 \small \textbf{Embodied Multi-Modal Agent (EMMA) trained by an LLM expert via cross-modality imitation learning.} EMMA takes a textual task instruction and pixel observations as its input state per step to generate a sequence of actions using a VLM. Then, we convert each pixel observation into a textual equivalent as the context of an LLM expert to produce improved actions for EMMA to imitate.}
    \label{fig:agent_archi}
% \vspace{-1.5em}
\end{figure*}

\section{Embodied Multi-Modal Agent}
Fig.~\ref{fig:agent_archi} illustrates the main idea of our ``Embodied Multi-Modal Agent (EMMA)'', whose detailed training procedures are given in Alg.~\ref{alg:training}. The agent is built upon a modularized VLM, which can follow instructions and interact with the environment through pixel observations and textual actions. To overcome these challenges associated with training EMMA, such as sparse reward or distribution shift, we explore the construction of an LLM expert from a parallel text world (Sec.~\ref{sec:llm_expert}) for providing EMMA with step-by-step guidance. Lastly, Sec.~\ref{sec:training} further discusses how to harness the LLM expert to train EMMA via cross-modality imitation learning. 
% In this section, we delineate the main components and training procedure underpinning EMMA, which are briefly illustrated in Fig.~\ref{fig:agent_archi}. And the detailed procedure is given in Alg.~\ref{alg:training}. Specifically, Sec.~\ref{sec:agent_archi} describes a modularized agent architecture that enables EMMA to interact with the visual world via text-based actions. Sec.~\ref{sec:llm_expert} introduces an LLM expert from a parallel TextWorld, which aims to provide EMMA with step-by-step guidance. Finally, Sec.~\ref{sec:training} discusses how to effectively harness the LLM expert to train EMMA via cross-modality imitation learning.

\subsection{EMMA in the Visual World}
\label{sec:agent_archi}
% [Agent's architecture]
In visual environments, EMMA $\pi_{\theta}$ is designed to process a task instruction $x_{\text{task}}$ (e.g., provided by human users) and the pixel observation $s_{v}^{t}$ at each interaction step $t$. It is expected to generate a sequence of high-level textual actions $\{x_{a}^{t}\sim\pi_{\theta}(\cdot|x_{\text{task}},s_{v}^{t})\}_{t=0}^{T}$\footnote{For brevity, we omit $x_{\text{task}}$ in the rest of this paper.} towards efficient completion of the task.
% In a visual environment with complicated scenes, the Emboided Multi-Modal Agent (EMMA) $\pi_{\theta}$ aims at taking a task description $x_{\text{task}}$ (e.g., the instruction from human users) and the pixel observation $s_{v}^{t}$ as its input state per step $t$ to produce a high-level text action $x_{a}^{t}=\pi_{\theta}(x_{\text{task}},s_{v}^{t})$\footnote{For simplicity, in the rest of this paper, we remove $x_{\text{task}}$.} for the efficient completion of the task. 
To achieve this, we draw inspiration from recent advances of large pretrained VLMs~\citep{corr/instructblip,icml/blip2,corr/minigpt4,corr/llava,corr/chunyuan_vlm_survey}, and modularize EMMA's architecture into four components: (1) a ViT to encode $s_{v}$ into visual embeddings, (2) a querying transformer (Q-Former) tailored to extract the most relevant visual features via the cross-attention between the visual embeddings and query tokens, (3) a linear projection layer to align visual features to text embeddings, (4) an LLM decoder taking the concatenation of the instruction tokens and the output of the linear projection layer to autoregressively generate the action $x_{a}$. 
In order to reduce computational overhead and prevent catastrophic forgetting~\citep{forgetting_in_llms}, we adopt the pretrained ViT, Q-Former, and LLM from InstructBLIP~\citep{corr/instructblip} and keep them frozen at the finetuning stage. We only update the linear projection layer, as illustrated in Fig.~\ref{fig:agent_archi}. Such a modularized architecture enables EMMA to integrate any existing pretrained vision models and LLMs in a flexible and computationally-efficient way.
% To minimize computation cost and catastrophic forgetting~\citep{forgetting_in_llms}, we adopt the pretrained ViT, Q-Former, and LLM from InstructBLIP~\citep{corr/instructblip} and keep them frozen during the training meanwhile only updating the parameters of the linear projection layer, as illustrated in Fig.~\ref{fig:agent_archi}. Such a modularized architecture enables EMMA to reuse any off-the-shelf pretrained vision model and LLM in a flexible and computationally-efficient way.

% [challenges] 
However, deploying EMMA into a complex visual world remains challenging. One of the main obstacles is that the direct use of any pretrained VLM is suboptimal because existing pretraining only focuses on static alignment between image-text pairs~\citep{icml/blip2,corr/instructblip,corr/chunyuan_vlm_survey}, so that the resulting agent may struggle to reason about the dynamics of the world.
As discussed in Sec.~\ref{sec:intro}, even the SOTA VLM, i.e., GPT-4V, fails to accomplish tasks in an embodied ALFWorld environment~\citep{iclr/alfworld}. In such a zero-shot setting, GPT-4V tends to rely on the linguistic prior of the currently detected objects, rather than the given task instruction and the underlying dynamics of the environment, to guide the interactions.
Moreover, finetuning a pretrained VLM on a pre-collected demonstration dataset is also suboptimal due to the diversity of environments and tasks~\citep{corr/rt-2}, the lack of large-scale expert annotations~\citep{open_x_embodiment_rt_x_2023} as well as the challenges posed by the distribution shift issue~\citep{corr/offlinerl_survey}.
A seemingly natural solution to the above challenges is reinforcement learning from environmental feedback (RLEF)~\citep{yang2023octopus}, in which reward signals rely on decomposing a task into a sequence of reasonable sub-goals and checking their completion. However, in real-world scenarios, most sub-goals cannot be defined or described precisely, so the reward is sparse; hence, we do not expect RLEF to be effective.

To this end, we propose to leverage interactive imitation learning (IL) to align EMMA with the dynamics of any environment, which however results in two critical algorithmic challenges: (1) How to obtain a high-quality, accessible, and scalable expert that EMMA can query during IL (Sec.~\ref{sec:llm_expert})? (2) Designing an effective strategy to train EMMA using this expert in complex, diverse, and potentially open-ended environments (Sec.~\ref{sec:training}).
% To circumvent the aforementioned challenges, we propose to leverage imitation learning (IL) to align EMMA with the dynamics of any environment, which however raises two algorithmic problems: (1) how to obtain a high-quality, accessible, and scalable expert queried by EMMA during IL (Sec.~\ref{sec:llm_expert}); (2) how to effectively train EMMA using the expert in a complex, diverse, and even open-ended environment (Sec.~\ref{sec:training}).

\subsection{LLM Expert from a Parallel TextWorld}
\label{sec:llm_expert}
% [How to build a parallel TextWorld and use it to unlock the potential of LLMs] 
Thanks to a series of prompting techniques in the realm of in-context learning, such as chain-of-thought~\citep{nips/cot}, tree-of-thought~\citep{yao2023tot}, ReAct~\citep{iclr/react}, and Reflexion~\citep{arxiv/reflexion}, pretrained LLMs have demonstrated impressive zero-shot performance across many decision-making scenarios~\citep{corr/drive_like_human,xu2023drivegpt4,icml/palm-e,rss/rt-1,corr/rt-2,huang2023voxposer,corr/embodiedgpt,icml/distill_vlm,arxiv/VIMA}. Despite the great potential in serving as high-quality and scalable experts, they are only able to interact with the environment via textual descriptions of the states, rather than using raw pixel observations like EMMA. 
To bridge this gap, we convert each pixel observation $s_{v}$ into a textual equivalent by extracting its metadata from the simulator~\citep{kolve2017ai2thor}, which is composed of attributes such as \texttt{Observed Objects}, \texttt{Observed Relations}, \texttt{Inventory}, and \texttt{Locations}. We then employ the Planning Domain Definition Language (PDDL)~\citep{1998pddl} to describe this metadata and create an equivalent textual description/state $s_{l}$ using the TextWorld engine~\citep{cote2019textworld}. Additional details are available in Appendix~\ref{appdix:tw_details} and an example of this process is illustrated in Fig.~\ref{fig:parallel_worlds}. This methodology enables the utilization of any pretrained LLM agent that generates a sequence of actions facilitating the training of EMMA through cross-modality IL between the two agents.
% To bridge this gap, we create text-based counterparts of each visual observation for LLM-based agents. Specifically, we extract metadata of the visual observation $s_{v}$, encompassing attributes such as \texttt{Observed Objects}, \texttt{Observed Relations}, \texttt{Inventory}, and \texttt{Location}, from the simulator~\citep{kolve2017ai2thor} and then use Planning Domain Definition Language (PDDL)~\citep{1998pddl} to describe the metadata and to construct a parallel text-based state $s_{l}$ using the TextWorld engine~\citep{cote2019textworld} (see Appendix [TBA] for additional details, and an example is provided in Fig.~\ref{fig:parallel_worlds}). In this way, we can harness a pretrained LLM agent (e.g., based on ChatGPT) to interact with the parallel TextWorld and produce a sequence of actions, which enables us to train EMMA via cross-modality IL between the two agents.

\subsection{Training EMMA via Cross-Modality Imitation}
\label{sec:training}

\begin{algorithm}[t!]
\caption{DAgger-DPO with a single task instruction}
\label{alg:training}
	\begin{algorithmic}[1]
    	\State \textbf{initialize:~}$i=0$, $\mathcal{D}\leftarrow\varnothing$, EMMA $\pi_{\theta}$, LLM actor $M_{a}$ with a FIFO memory pool $\mathcal{P}\leftarrow\varnothing$, LLM critic $M_{c}$
	    \State \textbf{input:~}max trials $I$, training epochs $I_{e}$, visual and textual envs $E_{v}$, $E_{l}$, instruction $x_{\text{task}}$, reference agent $\pi_{\text{ref}}$
            \State Initialize $\pi_{\theta}$ to $\pi_{\text{ref}}$ \Comment{\textcolor{mydarkred}{Behavior Cloning Initialization}}
            \While{task not completed or $i<I$}
                \State Get $\tau^{i}_{v}=[x_{\text{task}},s_{v}^{0},x_{a}^{0},\dots,s_{v}^{T},x_{a}^{T}]$ via $E_{v}$ with $\pi_{\theta}$
                \State Get $\tau^{i}_{l}=[x_{\text{task}},s_{l}^{0},x_{a}^{0},\dots,s_{l}^{T},x_{a}^{T}]$ via $E_{l}$ with $\tau^{i}_{v}$
                \State Generate retrospective feedback $\mathcal{P}_{i}=M_{c}(\tau^{i}_{l})$
                \State Update $\mathcal{P}$ with $\mathcal{P}_{i}$ (i.e., $\mathcal{P}\leftarrow\mathcal{P}\cup\mathcal{P}_{i}$)
                \For{$t=0$ to $T$} \Comment{\textcolor{mydarkred}{Dataset Aggregation}}
                    \State $x^{*}_{a}=M_{a}(\mathcal{P}, x_{\text{task}}, \dots, x_{a}^{t-1}, s_{l}^{t})$
                    \State $\mathcal{D}\leftarrow\mathcal{D}\cup\{(x_{\text{task}}, s_{v}^{t}, x_{a}^{t}, x^{*}_{a})\}$
                \EndFor
                \For{$j=0$ to $I_{e}-1$} \Comment{\textcolor{mydarkred}{Gradient Descent on $\theta$}}
                    \State Sample a mini-batch $\tau$ from $\mathcal{D}$
                    \State Update $\theta$ by minimizing Eq.~(\ref{eq:dpo}) with $\pi_{\text{ref}}$ and $\tau$
                \EndFor
                % \State Train $\pi_{\theta}$ on $\mathcal{D}$ by solving Eq.~(\ref{eq:dpo}) with $\pi_{\text{ref}}$
                \State $i\leftarrow i+1$
            \EndWhile
            \State \textbf{output:~}$\pi_{\theta^{*}}$
	\end{algorithmic}
\end{algorithm}

% [DAgger with DPO loss] 
Given an LLM expert from the parallel TextWorld, we aim to train a VLM agent $\pi_{\theta}$ in the visual world to imitate its behaviors closely. This is equal to minimizing the following objective under the distribution of states induced by $\pi_{\theta}$.
\begin{equation}
    \theta^{*} = \mathop{\arg\min}\limits_{\theta\in\Theta}\mathbb{E}_{\pi_{\theta}}\lft[\mathcal{L}_{\text{imit}}\lft(\pi_{\theta}(x_{a}|s_{v}),x^{*}_{a})\rgt)\rgt],
    \label{eq:imitate_obj}
\end{equation}
in which the choice of loss function $\mathcal{L}_{\text{imit}}$ is dependent on specific scenarios. For instance, it may be the expected cross-entropy loss for the discrete action space, or the expected MSE loss for the continuous action space. In our case, we select DPO~\citep{rafailov2024direct} loss due to its proven superior performance to the cross-entropy on aligning models with expert preferences within the discrete language space. Hence, Eq.~(\ref{eq:imitate_obj}) can be extended to the formulation below.
\begin{align}
    &\theta^{*} = \mathop{\arg\min}\limits_{\theta\in\Theta}-\mathbb{E}_{\pi_{\theta}}\lft[\mathcal{L}_{\text{imit}}\lft(\pi_{\theta},\pi_{\text{ref}},s_{v},x_{a},x^{*}_{a}\rgt)\rgt], \label{eq:dpo}\\ 
    &\mathcal{L}_{\text{imit}}(\cdot) \triangleq \log\sigma\lft(\beta\log\frac{\pi_{\theta}(x^{*}_{a}|s_{v})}{\pi_{\text{ref}}(x^{*}_{a}|s_{v})} - \beta\log\frac{\pi_{\theta}(x_{a}|s_{v})}{\pi_{\text{ref}}(x_{a}|s_{v})}\rgt) \nonumber
\end{align}
where $x^{*}_{a}$ is the action given by an expert while $x_{a}$ is the action given by $\pi_{\theta}$, $\sigma$ is the logistic function, and $\beta$ is a hyperparameter controlling the deviation from $\pi_{\text{ref}}$, i.e. the reference agent obtained by behavior cloning on a demonstration dataset produced by a rule-based expert in the visual world (see Appendix~\ref{appdix:dataset} for complete details). This regularization is essential, as it prevents the agent from deviating too far from the distribution on which the expert is accurate, as well as maintaining the generation diversity and avoiding premature convergence to some easy tasks~\citep{rafailov2024direct}. In practice, the VLM agent $\pi_{\theta}$ is also initialized to $\pi_{\text{ref}}$ for stabilizing the training process. Since environmental dynamics is both unknown and complex, we cannot compute the distribution of states visited by $\pi_{\theta}$ and can only sample it by rolling out the agent. Hence, Eq.~(\ref{eq:dpo}) is a non-i.i.d. supervised learning problem due to the dependence of the state distribution on $\pi_{\theta}$ itself, in which na\"ive behavior cloning faces issues like cumulative error and distribution shift~\citep{hussein2017imitation}. To address this, we employ an interactive IL algorithm, DAgger~\citep{aistats/dagger}, which provably converges to the optimal agent $\pi_{\theta^{*}}$.

% [Retrospective LLM expert] 
As discussed in Sec.~\ref{sec:llm_expert} and Sec.~\ref{sec:training}, we can harness an LLM expert to generate a sequence of actions, serving as $x^{*}_{a}$ in Eq.~(\ref{eq:dpo}). However, these actions may be suboptimal due to sparse environmental feedback~\citep{arxiv/reflexion,yang2023octopus} or defective in-context instructions~\citep{chen2023instructzero}. To collect better teaching signals, we introduce a ``retrospective LLM expert'' that is composed of two specialized models: an actor ($M_{a}$), built upon an API LLM and prompted to generate actions based on the task instruction and state observations; and a critic ($M_{c}$), also based on the same LLM, but designed to analyze EMMA's historical interactions and provide reflective feedback. A long-term memory $\mathcal{P}$ is maintained to store the feedback generated by $M_{c}$, which is then used to prompt $M_{a}$ for improved actions. The complete procedure is detailed in Line 7-10 of Alg.~\ref{alg:training}, and all prompts are provided in Sec.~\ref{appdix:prompts_llm} of the Appendix.

\section{Experiments}

\begin{figure*}[t!]
\begin{floatrow}
\ffigbox[0.33\textwidth]{%
\centering
\includegraphics[width=0.95\linewidth]{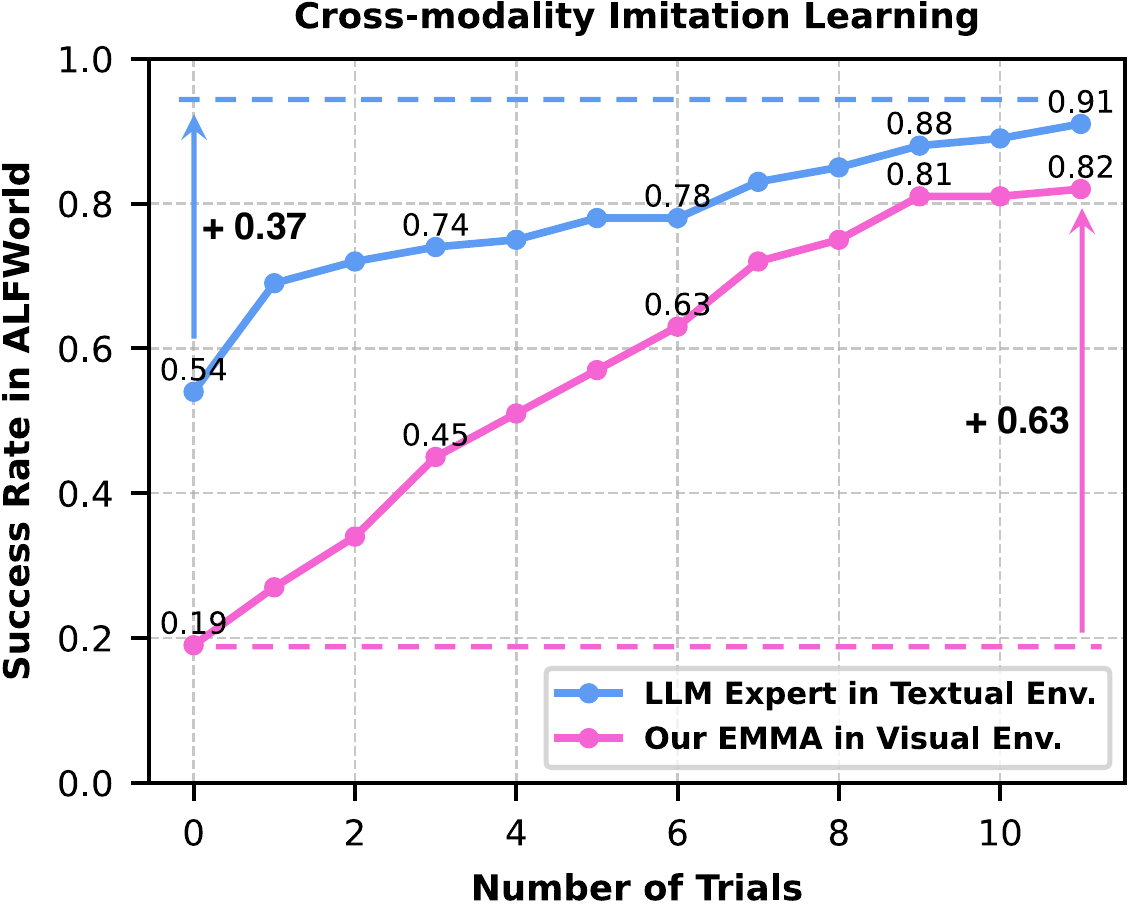}%
}{%
  \caption{\looseness=-1 \small \textbf{Comparison of success rate between EMMA and the retrospective LLM expert.} As the number of trials increases, the LLM expert progressively improves its performance through retrospective processes, and also the gap between the two agents decreases, indicating the effectiveness of cross-modality imitation. We separately plot the curve per task type in Fig.~\ref{fig:imitate_curve_sep} of Appendix.}
  \label{fig:imitate_curve}
}
\hspace{-15pt}
\capbtabbox[0.63\textwidth]{%
    \centering
    \renewcommand\arraystretch{1.1}
    \resizebox{\linewidth}{!}{
        \begin{tabular}{c@{\hspace{2.5pt}}l c@{\hspace{5pt}}c r@{\hspace{1.2pt}}lr@{\hspace{1.2pt}}lr@{\hspace{1.2pt}}lr@{\hspace{1.2pt}}lr@{\hspace{1.2pt}}lr@{\hspace{1.2pt}}lr@{\hspace{1.2pt}}l}
        \toprule
        \multicolumn{2}{l}{\multirow{2}{*}{\textbf{Agent}}} & \textbf{Visual} & \textbf{Textual} & \multicolumn{14}{c}{\textbf{Success Rate with Template Task Instruction}} \\
        % \hline
        \cline{5-18}
        % \textbf{Reference Vector}
        & & \textbf{Env.} & \textbf{Env.} & \multicolumn{2}{c}{Avg.} & \multicolumn{2}{c}{Pick} & \multicolumn{2}{c}{Clean} & \multicolumn{2}{c}{Heat} & \multicolumn{2}{c}{Cool} & \multicolumn{2}{c}{Look} & \multicolumn{2}{c}{Pick2} \\
        \midrule
        \multirow{2}{*}{\rotatebox[origin=c]{90}{\shortstack{VMs}}} 
        & \multicolumn{1}{l}{ResNet-18*~\citep{iclr/alfworld}} &\cmark &\xmark &0.06&{\color[HTML]{525252}(-)} &\multicolumn{2}{c}{\cellcolor{blank}-} &\multicolumn{2}{c}{\cellcolor{blank}-} &\multicolumn{2}{c}{\cellcolor{blank}-} &\multicolumn{2}{c}{\cellcolor{blank}-} &\multicolumn{2}{c}{\cellcolor{blank}-} &\multicolumn{2}{c}{\cellcolor{blank}-} \\
        & \multicolumn{1}{l}{MCNN-FPN*~\citep{iclr/alfworld}} &\cmark &\xmark &0.05&{\color[HTML]{525252}(-)} &\multicolumn{2}{c}{\cellcolor{blank}-} &\multicolumn{2}{c}{\cellcolor{blank}-} &\multicolumn{2}{c}{\cellcolor{blank}-} &\multicolumn{2}{c}{\cellcolor{blank}-} &\multicolumn{2}{c}{\cellcolor{blank}-} &\multicolumn{2}{c}{\cellcolor{blank}-} \\
        \midrule
        \multirow{6}{*}{\rotatebox[origin=c]{90}{\shortstack{LMs}}} 
        & \multicolumn{1}{l}{BUTLER*~\citep{iclr/alfworld}} &\xmark &\cmark &0.26&{\color[HTML]{525252}(-)} &0.31&{\color[HTML]{525252}(-)} &0.41&{\color[HTML]{525252}(-)} &0.60&{\color[HTML]{525252}(-)} &0.27&{\color[HTML]{525252}(-)} &0.12&{\color[HTML]{525252}(-)} &0.29&{\color[HTML]{525252}(-)} \\
        & \multicolumn{1}{l}{GPT-BUTLER~\citep{emnlp/gpt-butler}} &\xmark &\cmark &0.69&{\color[HTML]{525252}(18.8)} &0.62&{\color[HTML]{525252}(18.4)} &0.81&{\color[HTML]{525252}(18.4)} &\textbf{0.85}&{\color[HTML]{525252}(12.7)} &0.78&{\color[HTML]{525252}(15.6)} &0.50&{\color[HTML]{525252}(24.4)} &0.47&{\color[HTML]{525252}(26.6)} \\
        & \multicolumn{1}{l}{ReAct~\citep{iclr/react}} &\xmark &\cmark &0.54&{\color[HTML]{525252}(20.6)} &0.71&{\color[HTML]{525252}(18.1)} &0.65&{\color[HTML]{525252}(18.8)} &0.62&{\color[HTML]{525252}(18.2)} &0.44&{\color[HTML]{525252}(23.2)} &0.28&{\color[HTML]{525252}(23.7)} &0.35&{\color[HTML]{525252}(25.5)} \\
        & \multicolumn{1}{l}{Reflexion~\citep{arxiv/reflexion}} &\xmark &\cmark &\textbf{0.91}&{\color[HTML]{525252}(18.7)} &\textbf{0.96}&{\color[HTML]{525252}(17.4)} &\textbf{1.00}&{\color[HTML]{525252}(17.0)} &0.81&{\color[HTML]{525252}(19.4)} &0.83&{\color[HTML]{525252}(21.6)} &0.94&{\color[HTML]{525252}(16.9)} &\textbf{0.88}&{\color[HTML]{525252}(21.6)} \\
        & \multicolumn{1}{l}{DEPS*~\citep{wang2023deps}} &\xmark &\cmark &0.76&{\color[HTML]{525252}(-)} &0.93&{\color[HTML]{525252}(-)} &0.50&{\color[HTML]{525252}(-)} &0.80&{\color[HTML]{525252}(-)} &\textbf{1.00}&{\color[HTML]{525252}(-)} &\textbf{1.00}&{\color[HTML]{525252}(-)} &0.00&{\color[HTML]{525252}(-)} \\
        & \multicolumn{1}{l}{AutoGen*~\citep{wu2023autogen}} &\xmark &\cmark &0.77&{\color[HTML]{525252}(-)} &0.92&{\color[HTML]{525252}(-)} &0.74&{\color[HTML]{525252}(-)} &0.78&{\color[HTML]{525252}(-)} &0.86&{\color[HTML]{525252}(-)} &0.83&{\color[HTML]{525252}(-)} &0.41&{\color[HTML]{525252}(-)} \\
        \midrule
        \multirow{5}{*}{\rotatebox[origin=c]{90}{\shortstack{VLMs}}} 
        & \multicolumn{1}{l}{MiniGPT-4~\citep{corr/minigpt4}} &\cmark &\xmark &0.16&{\color[HTML]{525252}(26.9)} &0.04&{\color[HTML]{525252}(29.0)} &0.00&{\color[HTML]{525252}(30.0)} &0.19&{\color[HTML]{525252}(26.3)} &0.17&{\color[HTML]{525252}(26.7)} &0.67&{\color[HTML]{525252}(17.7)} &0.06&{\color[HTML]{525252}(28.9)} \\
        & \multicolumn{1}{l}{BLIP-2~\citep{icml/blip2}} &\cmark &\xmark &0.04&{\color[HTML]{525252}(29.5)} &0.00&{\color[HTML]{525252}(30.0)} &0.06&{\color[HTML]{525252}(29.3)} &0.04&{\color[HTML]{525252}(29.9)} &0.11&{\color[HTML]{525252}(28.2)} &0.06&{\color[HTML]{525252}(29.2)} &0.00&{\color[HTML]{525252}(30.0)} \\
        & \multicolumn{1}{l}{LLaMA-Adapter~\citep{corr/llama-adaptor}} &\cmark &\xmark &0.13&{\color[HTML]{525252}(27.5)} &0.17&{\color[HTML]{525252}(26.4)} &0.10&{\color[HTML]{525252}(28.6)} &0.27&{\color[HTML]{525252}(24.2)} &0.22&{\color[HTML]{525252}(26.7)} &0.00&{\color[HTML]{525252}(30.0)} &0.00&{\color[HTML]{525252}(30.0)} \\
        & \multicolumn{1}{l}{InstructBLIP~\citep{corr/instructblip}} &\cmark &\xmark &0.22&{\color[HTML]{525252}(26.2)} &0.50&{\color[HTML]{525252}(21.5)} &0.26&{\color[HTML]{525252}(25.0)} &0.23&{\color[HTML]{525252}(27.2)} &0.06&{\color[HTML]{525252}(28.9)} &0.17&{\color[HTML]{525252}(26.8)} &0.00 &{\color[HTML]{525252}(30.0)} \\
        & \multicolumn{1}{l}{\textbf{EMMA} (Ours)} &\cmark &\xmark &0.82&{\color[HTML]{525252}(19.5)} &\textbf{0.71}&{\color[HTML]{525252}(19.3)} &\textbf{0.94}&{\color[HTML]{525252}(17.5)} &\textbf{0.85}&{\color[HTML]{525252}(19.6)} &\textbf{0.83}&{\color[HTML]{525252}(19.9)} &\textbf{0.88}&{\color[HTML]{525252}(19.6)} &\textbf{0.67}&{\color[HTML]{525252}(22.4)} \\
        \midrule
        \multicolumn{2}{l}{\textbf{Human Performance*}~\citep{cvpr/alfred}} &\cmark &\xmark &\textbf{0.91}&{\color[HTML]{525252}(-)} &\multicolumn{2}{c}{\cellcolor{blank}-} &\multicolumn{2}{c}{\cellcolor{blank}-} &\multicolumn{2}{c}{\cellcolor{blank}-} &\multicolumn{2}{c}{\cellcolor{blank}-} &\multicolumn{2}{c}{\cellcolor{blank}-} &\multicolumn{2}{c}{\cellcolor{blank}-} \\
        \bottomrule 
        \end{tabular}
    }
}{%
  \caption{\looseness=-1 \small \textbf{Comparison with the state of the arts.} $*$-reported in previous work. VMs = vision models, LMs = language models, VLMs = vision-language models. ``Visual Env.'' and ``Textual Env.'' refer to the visual and the parallel textual environments from ALFWorld~\citep{iclr/alfworld}, respectively. \text{\cmark/\xmark} denotes the corresponding environment used/not used to deploy the agent. The highest scores for each task in the same type of environment are highlighted in \textbf{bold}. The average interaction steps are given in the \text{\color[HTML]{525252}(parentheses)}. \textbf{EMMA substantially outperforms other SOTA VLM agents in the visual environments, and its success also directs a promising way to achieve human-level performance in ALFWorld.}}
  \label{tab:main_results}
}
\end{floatrow}
% \vspace{-10pt}
\end{figure*}

\subsection{Experimental Setup}
\label{sec:exp_setup}
\textbf{Environments.}
We base our experiments on the ALFWorld benchmark~\citep{iclr/alfworld}, a cross-modality simulation platform that encompasses a wide range of embodied household tasks. For each task, ALFWorld integrates a visual environment, rendered by the Ai2Thor simulator~\citep{kolve2017ai2thor}, with a corresponding textual environment. Textual environments employ the Planning Domain Definition Language (PDDL)~\citep{1998pddl} to translate each pixel observation from the simulator into an equivalent text-based observation, and then construct an interactive world using the TextWorld engine~\citep{cote2019textworld}. Fig.~\ref{fig:parallel_worlds} provides an illustrative example of the tasks featured in ALFWorld.
The tasks within the benchmark are categorized into 6 types: Pick \& Place, Clean \& Place, Heat \& Place, Cool \& Place, Look in Light, and Pick Two Objects \& Place. Each task requires an agent to execute a series of text-based actions, such as ``go to safe 1'', ``open safe 1'', or ``heat egg 1 with microwave 1'', following a predefined instruction. These actions involve navigating and interacting with the environment. To provide a comprehensive understanding, we have visualized an example of each task type in Fig.~\ref{fig:full_example_tasks} of the Appendix.
A task in this benchmark may involve interactions with over 10 objects and require more than 30 steps for a human expert to solve, thus challenging an agent's capabilities in long-horizon planning, instruction following, and the utilization of commonsense knowledge. 
For a fair comparison, we follow the same setting as prior work~\citep{cvpr/alfred,iclr/alfworld,emnlp/gpt-butler,iclr/react,arxiv/reflexion} and evaluate all baselines using 134 out-of-distribution (OOD) tasks. \looseness -1

\noindent\textbf{Baselines.}
To verify the effectiveness of cross-modality imitation learning, we compare our EMMA with several baselines and SOTA agents using the ALFWorld benchmark with both visual and textual environments. 
The compared agents can be divided into three categories: vision models, language models, and vision-language models. 
Concretely, vision models, including ResNet-18~\citep{cvpr/He} and MCNN-FPN~\citep{iccv/mask-rcnn}, utilize pretrained vision encoders to extract salient features from each pixel observation. The extracted features then serve as input for a Multi-Layer Perceptron (MLP) policy, which is trained by behavior cloning on a pre-collected demonstration dataset. 
Unlike vision models performing in the visual environment, language models complete the same tasks but in a parallel, text-based environment. BUTLER~\citep{iclr/alfworld} employs a transformer seq2seq model enhanced with a pointer softmax mechanism~\citep{acl/GulcehreANZB16}. This architecture aggregates previous observations as input to generate text-based actions token-by-token. GPT-BUTLER~\citep{emnlp/gpt-butler}, a variant of the GPT-2 model~\citep{radford2019gpt2}, is initially pretrained on a static demonstration dataset and further finetuned using data collected online. ReAct~\cite{iclr/react} takes a novel approach by utilizing LLMs to generate reasoning traces and task-specific actions in an interleaved manner. This method aids the agent in developing, tracking, and updating its action plans interactively. Reflexion~\citep{arxiv/reflexion} similarly employs an LLM, but it focuses on reflecting upon environmental feedback. It maintains this reflective text in an episodic memory buffer, enhancing the agent's ability to improve actions in subsequent trials. Similar to Reflexion, a concurrent work, DEPS~\citep{wang2023deps}, also corrects errors in previous LLM-generated actions by integrating descriptions of the action execution process and providing self-explanations for the feedback. Moreover, beyond the single-agent framework, AutoGen~\citep{wu2023autogen} exhibits the potential to accomplish a broad spectrum of tasks through the cooperation of multiple LLM agents.
Finally, we consider a range of vision-language models, such as MiniGPT-4~\citep{corr/minigpt4}, BLIP-2~\citep{icml/blip2}, LLaMA-Adaptor~\citep{corr/llama-adaptor}, and InstructBLIP~\citep{corr/instructblip}, as agents to interact with the visual environment. Unlike pure vision or language models, VLMs are designed to process and integrate both visual and textual data, offering a more holistic understanding of the environment. To align these agents with the specific requirements of the ALFWorld benchmark, we finetune them on a pre-collected demonstration dataset. This finetuning process is crucial as it enables the agent to comprehend and adhere to ALFWorld's unique grammar and to develop a basic ``gamesense''. 

\subsection{Training Details}
\label{sec:training_detail}
The architectural design of EMMA is depicted in Fig.~\ref{fig:agent_archi}. At its core, EMMA employs a Query Transformer (Q-Former) to process visual data. This Q-Former extracts features using a frozen ViT encoder. Its output consists of 32 visual tokens, which are then passed through a linear projection layer before being fed to a frozen LLM decoder. Similar to other VLM agents, EMMA is also finetuned on a pre-collected demonstration dataset, aligning its basic ability with the ALFWorld benchmark.
In order to align EMMA with the dynamics of ALFWorld, we train it by imitating an LLM expert (see Alg.~\ref{alg:training}). We choose \textit{text-davinci-003} developed by OpenAI as our LLM expert because of its established capabilities in reasoning and planning~\citep{arxiv/reflexion,iclr/react,wu2023autogen,wang2023deps}. In this setup, \textit{text-davinci-003} serves dual roles: it serves as an actor, providing EMMA with expert actions, and as a critic, analyzing EMMA's historical trajectories. This analysis generates retrospective feedback, which is then incorporated into the actor's long-term memory, leading to improved actions in future trials. Further details about the hyperparameters and prompts used in our training procedure are available in Table~\ref{tab:hyperparameters} of the Appendix. \looseness -1

\subsection{Comparison with State of the Art}
\label{sec:main_results}

\noindent\textbf{EMMA sets new SOTA performance in visual ALFWorld.}
In this section, we compare EMMA with 12 other representative agents on the ALFWorld benchmark, spanning both visual and textual environments (refer to Table~\ref{tab:main_results}). We assess two key metrics: the success rate, which is the percentage of trials completed successfully, and the average number of interaction steps required for task completion, with a lower number indicating higher efficiency. EMMA demonstrates superior performance in both metrics, significantly outperforming all VLM agents in visual environments. This achievement underscores the effectiveness of our cross-modality imitation learning approach, as depicted in the learning curve shown in Fig.~\ref{fig:imitate_curve}.
Furthermore, EMMA's performance markedly exceeds that of VM agents, highlighting the crucial role of the prior knowledge embedded in VLMs.
Intriguingly, EMMA's performance is comparable with LLM agents that operate using perfectly semantic descriptions of visual observations. This is largely attributed to EMMA's training strategy of imitating an expert LLM agent, proving to be more efficient than learning from scratch in a purely visual setting. As a result, EMMA stands out as the only VLM agent that substantially surpasses SOTA LLM agents, such as AutoGen~\citep{wu2023autogen} and DEPS~\citep{wang2023deps}, in these environments. And its success also directs a potential way to achieve human-level performance in the visual environments of ALFWorld.

\begin{figure}[t!]
% \vspace{-0.2em}
    \centering
    \includegraphics[width=0.9\linewidth]{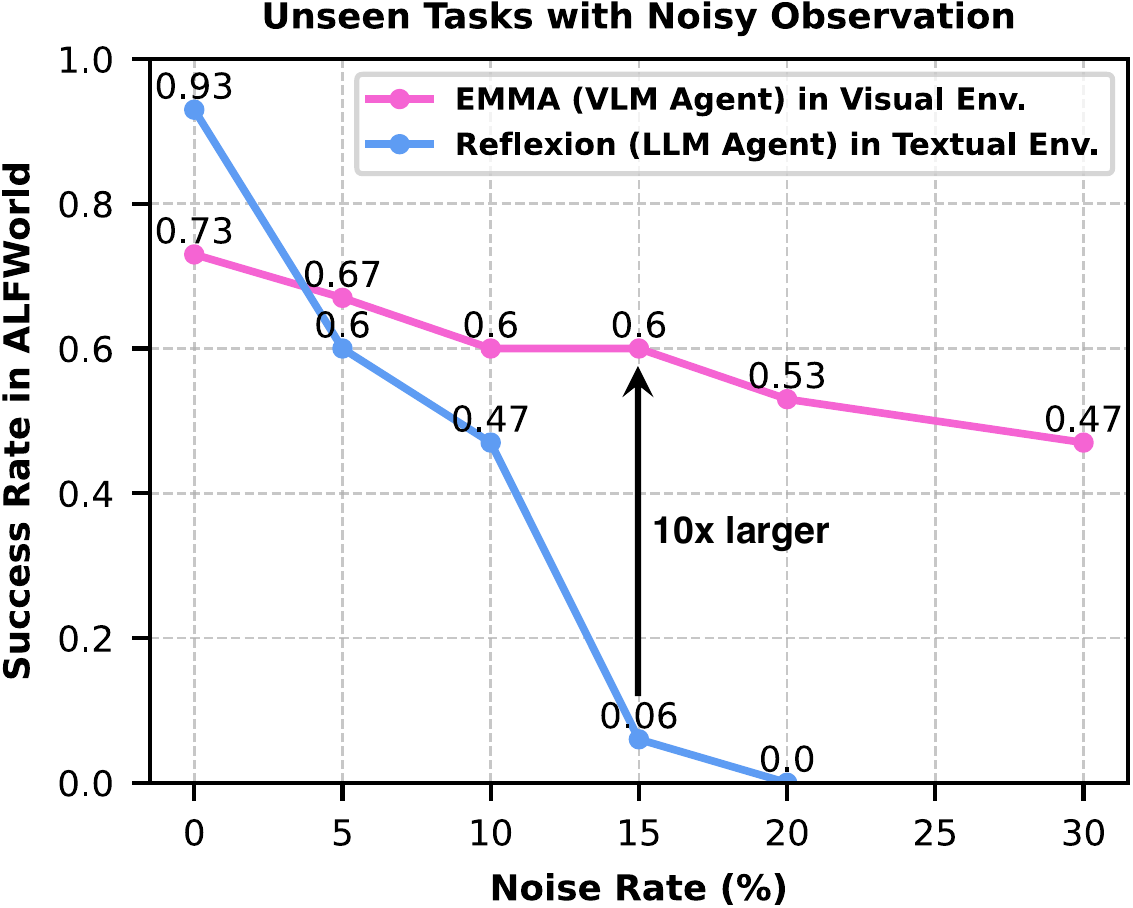}
    % \vspace{-0.3em}
    \caption{\looseness=-1 \small \textbf{Comparison of robustness between VLM and LLM agents.} Given a specific noise rate, we crop a random portion of the pixel observation and resize it to a given size as the new observation. Similarly, we randomly replace some tokens in the textual observation with arbitrary ones.}
    \label{fig:robust_comparison}
%     \vspace{-0.4cm}
\end{figure}

\noindent\textbf{EMMA is more robust to noisy observations than LLM agents.}
While LLM agents exhibit a higher success rate with fewer interaction steps in textual environments, as indicated in Table~\ref{tab:main_results}, we hypothesize that this superior performance largely relies on their precise semantic abstraction of the environment. However, such an abstraction might not be feasible in real-world applications. To verify this assumption, we set up a more practical scenario where observations are deliberately perturbed at a specific noise rate. We then compare the robustness of EMMA and a SOTA LLM agent, Reflexion, under these noisy observations.
To generate noisy observations, a random portion of the visual observation is cropped, resized, and then used to replace the original observation. Similarly, in the textual observation, random tokens are substituted with arbitrary ones. 
As illustrated in Fig.~\ref{fig:robust_comparison}, with the noise rate increasing, EMMA's performance remains significantly more robust than Reflexion. This could be attributed to the vision encoder in the VLM, which is adept at filtering out visual noises. On the other hand, textual noises are directly processed by the LLM, which can substantially impair the performance of LLM-based agents. This finding highlights the potential advantages of VLM agents like EMMA in visual scenarios, in which data is often imperfect and noisy.

% While LLM agents achieve a higher success rate using lower interaction steps in textual environments, as shown in Table~\ref{tab:main_results}, we assume that their superior performance primarily benefits from the precisely semantic abstraction of the environment, which is usually impractical for real-world applications. To verify this hypothesis, we consider a more practical setting, in which observations are perturbed given a specific noise rate, and make a comparison of robustness between EMMA and a SOTA LLM agent (i.e., Reflexion). Concretely, we crop a random portion of the visual observation, resize it to a given size, and then replace the original observation with the cropped one. Similarly, we randomly replace some tokens in the textual observation with arbitrary ones. As demonstrated in Fig.~\ref{fig:robust_comparison}, with the increasing noise rate, we find that the performance of EMMA is much more robust than Reflexion. One possible explanation is that the vision encoder of the VLM can easily filter out the visual noises. In contrast, the textual noises are directly propagated to the LLM and thus significantly degrade the performance of LLM-based agents.

\subsection{Ablation Study}
\label{sec:ablation}
\begin{figure*}[htb]
\centering
% \vspace{-0.3em}
    \small
        {\includegraphics[width=0.32\textwidth]{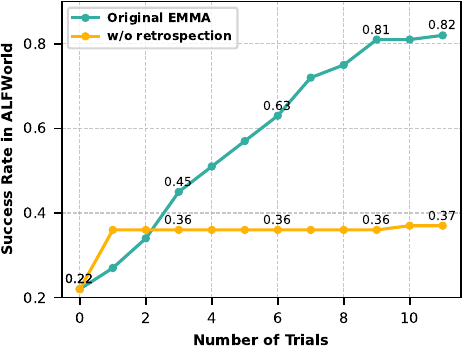}
        % \label{fig:ab_reflect}
        }
        {\includegraphics[width=0.32\textwidth]{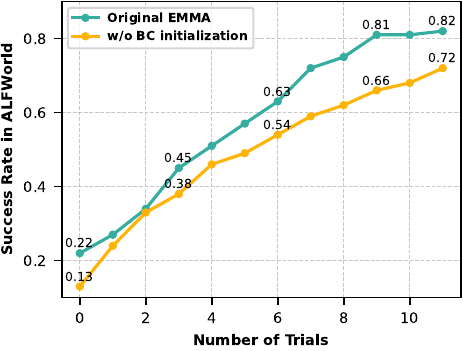}
        % \label{fig:ab_ref}
        }
        {\includegraphics[width=0.32\textwidth]{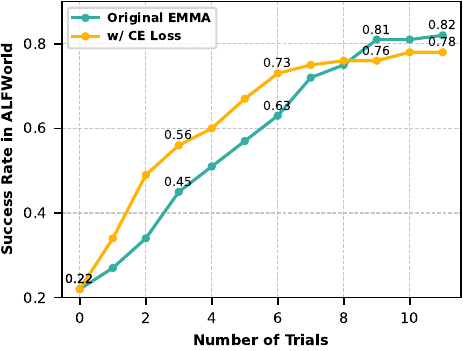}
        % \label{fig:ab_loss}
        }
    % \vspace{-1em}
    % \vspace{-5pt}
    \caption{\textbf{Ablation study.} \textit{left}: The ability of LLM expert to reflect on and learn from past actions appears to be a key factor in achieving the impressive results we observed. \textit{middle}: The ``w/o BC initialization'' variant uses a pretrained VLM instead of the reference agent $\pi_{\text{ref}}$ to initialize EMMA. \textit{right}: The 'w/ CE Loss' variant replaces the DPO loss (described in Eq.~\ref{eq:dpo}) with a token-level cross-entropy (CE) loss. Please refer to Sec.~\ref{sec:ablation} for a thorough discussion.}
    \label{fig:ablation}
% \vspace{-1em}
\end{figure*}

\textbf{Retrospection improves EMMA over time.}
To assess the importance of the retrospective LLM expert, we present the average success rate of EMMA after each trial and compare it with a key variation: ``EMMA w/o Retrospection'', as shown in Fig.~\ref{fig:ablation} (left). This variant of EMMA is trained using the same procedure as the original EMMA but removes the retrospective process. Instead, it relies solely on a plain LLM actor to provide relabeled actions.
The results show that EMMA with the retrospective mechanism significantly outperforms its counterpart. This finding is crucial as it indicates that the retrospective process is not just a supplementary feature but a fundamental component of EMMA's architecture that contributes substantially to its enhanced performance. \looseness-1
% In essence, the ability of EMMA to reflect on and learn from past actions appears to be a key factor in achieving the impressive results we observed.

\noindent\textbf{EMMA benefits from BC initialization.}
Through an ablation study, we evaluate the impact of behavior cloning (BC) initialization, a process described in line 3 of Alg.~\ref{alg:training}. 
Fig.~\ref{fig:ablation} (middle) demonstrates that EMMA, when deprived of BC initialization, experiences a slight reduction in the average success rate across 134 unseen tasks compared to its original setup. Despite this decrease, EMMA without BC initialization still outperforms other VLM agents, as clearly shown when compared with the results in Table~\ref{tab:main_results}.
Furthermore, Fig.~\ref{fig:ablation_bc_separate} in the Appendix breaks down the result by task type. It reveals a consistent but slight drop in performance across 5 out of the 6 task types. These results reflect that while BC initialization contributes positively to EMMA's overall performance, it is not critical for achieving the notable results we have reported.
% We evaluate the impact of behavior cloning (BC) initialization (detailed in line 3 of Alg.~\ref{alg:training}) by conducting an ablation study. Fig.~\ref{fig:ab_ref} illustrates that EMMA, without BC initialization, exhibits a slight decrease in the average success rate across 134 unseen tasks compared to the original configuration. However, it still significantly outperforms other VLM agents, as evident from the comparison with Table~\ref{tab:main_results}. Additionally, Fig.~\ref{fig:ablation_bc_separate} of the Appendix presents the results by task type, revealing a consistently slight decline in performance over 5 out of 6 task types. These findings suggest that BC initialization can enhance EMMA's performance but is not crucial for achieving the notable results we report.

\noindent\textbf{DPO enables more effective imitation learning.}
To evaluate the effectiveness of using DPO loss in Eq.~(\ref{eq:imitate_obj}), we conducted an ablation study with an alternative version of EMMA, referred to ``EMMA w/ Cross Entropy (CE) Loss''. In this variant, EMMA is optimized using the token-level CE loss, a common objective for finetuning VLMs.
The results, as depicted in Fig.~\ref{fig:ablation} (right), reveal that EMMA w/ CE Loss does not achieve the same high success rate as the original EMMA with DPO loss, suggesting that DPO loss contributes to enhancing the upper performance bound of EMMA.
In addition, we noted that EMMA w/ CE Loss exhibits faster convergence in the initial training stages than the original EMMA. This premature convergence leads to the agent paying attention to the expert actions from easier tasks, usually addressed in the first few training epochs. This can suppress EMMA's exploration and learning on more complex tasks, resulting in worse performance.
% To assess the effectiveness of using DPO loss in Eq.~\ref{eq:imitate_obj}, we study an ablation, EMMA w/ Cross Entropy (CE) Loss, which directly optimizes EMMA with the token-level CE loss, a widely used objective to finetune VLMs, on the expert actions. The results in Fig.~\ref{fig:ablation} (right) show that the final success rate of EMMA w/ CE Loss is inferior to the original EMMA, indicating that the DPO further improves the upper bound of EMMA's performance. We also observe that the EMMA w/ CE Loss achieves faster convergence than the original EMMA in the early stage of training, which is mainly due to the fact that EMMA with the CE loss tends to fit the expert actions from easier tasks that are usually solved in the first few training epochs. This characteristic suppresses the exploration of EMMA on more complex tasks.

\vspace{-0.1em}
\subsection{Generalization to Free-form Task Instructions}
\vspace{-0.1em}
\begin{figure*}[t]
% \vspace{-5pt}
\setlength{\tabcolsep}{5.5pt}
\begin{minipage}{0.7\textwidth}
\centering
\renewcommand\arraystretch{1.1}
\resizebox{\textwidth}{!}{
    \begin{tabular}{c@{\hspace{2.5pt}}l c@{\hspace{5pt}}c r@{\hspace{1.2pt}}lr@{\hspace{1.2pt}}lr@{\hspace{1.2pt}}lr@{\hspace{1.2pt}}lr@{\hspace{1.2pt}}lr@{\hspace{1.2pt}}lr@{\hspace{1.2pt}}l}
        \toprule
        \multicolumn{2}{l}{\multirow{2}{*}{\textbf{Agent}}} & \textbf{Visual} & \textbf{Textual} & \multicolumn{14}{c}{\textbf{Success Rate with Free-form Task Instruction}} \\
        % \hline
        \cline{5-18}
        % \textbf{Reference Vector}
        & & \textbf{Env.} & \textbf{Env.} & \multicolumn{2}{c}{Avg.} & \multicolumn{2}{c}{Pick} & \multicolumn{2}{c}{Clean} & \multicolumn{2}{c}{Heat} & \multicolumn{2}{c}{Cool} & \multicolumn{2}{c}{Look} & \multicolumn{2}{c}{Pick2} \\
        \midrule
        \multirow{4}{*}{\rotatebox[origin=c]{90}{\shortstack{LMs}}} 
        & \multicolumn{1}{l}{BUTLER*~\citep{iclr/alfworld}} &\xmark &\cmark &0.03&{\color[HTML]{525252}(-)} &0.10&{\color[HTML]{525252}(-)} &0.22&{\color[HTML]{525252}(-)} &0.05&{\color[HTML]{525252}(-)} &0.17&{\color[HTML]{525252}(-)} &0.00&{\color[HTML]{525252}(-)} &0.00&{\color[HTML]{525252}(-)} \\
        & \multicolumn{1}{l}{GPT-BUTLER~\citep{emnlp/gpt-butler}} &\xmark &\cmark &0.31&{\color[HTML]{525252}(24.7)} &0.25&{\color[HTML]{525252}(25.4)} &0.42&{\color[HTML]{525252}(23.5)} &0.46&{\color[HTML]{525252}(20.9)} &0.44&{\color[HTML]{525252}(21.8)} &0.00&{\color[HTML]{525252}(30.0)} &0.12&{\color[HTML]{525252}(28.9)} \\
        & \multicolumn{1}{l}{ReAct~\citep{iclr/react}} &\xmark &\cmark &0.37&{\color[HTML]{525252}(23.6)} &0.46&{\color[HTML]{525252}(22.0)} &0.39&{\color[HTML]{525252}(23.8)} &0.50&{\color[HTML]{525252}(20.5)} &0.28&{\color[HTML]{525252}(24.6)} &0.22&{\color[HTML]{525252}(26.4)} &0.29&{\color[HTML]{525252}(25.9)} \\
        & \multicolumn{1}{l}{Reflexion~\citep{arxiv/reflexion}} &\xmark &\cmark &\textbf{0.78}&{\color[HTML]{525252}(17.0)} &\textbf{0.84}&{\color[HTML]{525252}(15.8)} &\textbf{0.74}&{\color[HTML]{525252}(18.4)} &\textbf{0.77}&{\color[HTML]{525252}(17.2)} &\textbf{0.61}&{\color[HTML]{525252}(19.4)} &\textbf{0.94}&{\color[HTML]{525252}(13.8)} &\textbf{0.82}&{\color[HTML]{525252}(16.7)} \\
        \midrule
        \multirow{5}{*}{\rotatebox[origin=c]{90}{\shortstack{VLMs}}} 
        & \multicolumn{1}{l}{MiniGPT-4~\citep{corr/minigpt4}} &\cmark &\xmark &0.00&{\color[HTML]{525252}(30.0)} &0.00&{\color[HTML]{525252}(30.0)} &0.00&{\color[HTML]{525252}(30.0)} &0.00&{\color[HTML]{525252}(30.0)} &0.00&{\color[HTML]{525252}(30.0)} &0.00&{\color[HTML]{525252}(30.0)} &0.00&{\color[HTML]{525252}(30.0)} \\
        & \multicolumn{1}{l}{BLIP-2~\citep{icml/blip2}} &\cmark &\xmark &0.01&{\color[HTML]{525252}(29.7)} &0.04&{\color[HTML]{525252}(29.1)} &0.03&{\color[HTML]{525252}(29.5)} &0.00&{\color[HTML]{525252}(30.0)} &0.00&{\color[HTML]{525252}(30.0)} &0.00&{\color[HTML]{525252}(30.0)} &0.00&{\color[HTML]{525252}(30.0)} \\
        & \multicolumn{1}{l}{LLaMA-Adapter~\citep{corr/llama-adaptor}} &\cmark &\xmark &0.02&{\color[HTML]{525252}(29.6)} &0.04&{\color[HTML]{525252}(29.3)} &0.03&{\color[HTML]{525252}(29.4)} &0.04&{\color[HTML]{525252}(29.3)} &0.00&{\color[HTML]{525252}(30.0)} &0.00&{\color[HTML]{525252}(30.0)} &0.00&{\color[HTML]{525252}(30.0)} \\
        & \multicolumn{1}{l}{InstructBLIP~\citep{corr/instructblip}} &\cmark &\xmark &0.01&{\color[HTML]{525252}(29.8)} &0.00&{\color[HTML]{525252}(30.0)} &0.03&{\color[HTML]{525252}(29.3)} &0.00&{\color[HTML]{525252}(30.0)} &0.00&{\color[HTML]{525252}(30.0)} &0.00&{\color[HTML]{525252}(30.0)} &0.00&{\color[HTML]{525252}(30.0)} \\
        & \multicolumn{1}{l}{\textbf{EMMA} (Ours)} &\cmark &\xmark &\textbf{0.68}&{\color[HTML]{525252}(22.0)} &\textbf{0.72}&{\color[HTML]{525252}(21.7)} &\textbf{0.65}&{\color[HTML]{525252}(22.7)} &\textbf{0.72}&{\color[HTML]{525252}(22.0)} &\textbf{0.56}&{\color[HTML]{525252}(23.9)} &\textbf{0.67}&{\color[HTML]{525252}(23.3)} &\textbf{0.76}&{\color[HTML]{525252}(18.1)} \\
        \bottomrule 
    \end{tabular}
}
\end{minipage}
\hfill
\begin{minipage}{0.28\textwidth} %
    \centering
    \includegraphics[width=\textwidth]{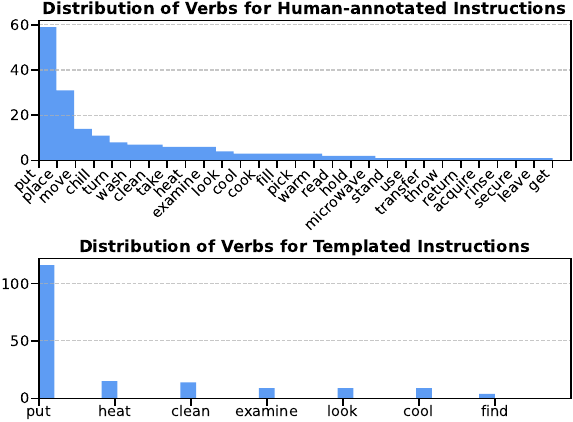}
\end{minipage}
% \vspace{-5pt}
\caption{\looseness=-1 \small \textbf{Generalization to open-vocabulary, free-form task instructions.} \textit{Left}: These instructions are provided by different human annotators using Amazon Mechanical Turk~\citep{cvpr/alfred}. \textit{Right}: Frequency distribution of verbs for human-annotated and templated task instructions.}
\label{tab:generalization}
%\vspace{-0.2cm}
\end{figure*}

The ability of AI agents to preciously follow task instructions given in open-vocabulary and free-form text is crucial for real-world problems. To assess this, we conducted an additional experiment focusing on the generalization capabilities of EMMA, an agent trained with template task instructions.
In this experiment, we re-evaluated the performance of EMMA and other baseline agents on 134 unseen tasks, using human-annotated instructions instead of template ones. These human-annotated instructions include a large amount of OOD verbs and objects, presenting a more realistic and challenging scenario for the agents to address. To underscore this challenge, we compared the vocabulary distribution between the template and human-annotated instructions, as shown in Fig.~\ref{tab:generalization} (right). Moreover, we provide a comprehensive analysis of the vocabulary used across both instruction types in Fig.~\ref{fig:all_words_distribution} of the Appendix.
% The proficiency of the trained agents in following user-provided task instructions, usually those posed with open-vocabulary and free-form text, is pivotal for their practical applicability. Hence, we conducted an additional experiment to investigate the generalization of EMMA, the agent trained with templated task instructions, to human-annotated instructions. We re-evaluate the performance of EMMA and other baselines on 134 unseen tasks. Instead of templated instructions in Sec.~\ref{sec:main_results}, we adopt human-annotated task instructions encompassing numerous OOD verbs and objects. The comparison of vocabulary distribution between templated and human-annotated instructions is depicted in Fig.~\ref{tab:generalization} (right). A comprehensive analysis is provided in Fig.~\ref{fig:all_words_distribution} of the Appendix.
In Fig.~\ref{tab:generalization} (left), EMMA demonstrates a slight performance decline, while a significant degradation is observed in other baselines. We also note that Reflexion, a SOTA LLM agent, exhibits exceptional generalization to those OOD instructions.
According to these empirical results in Fig.~\ref{tab:generalization}, we have the following conclusions: (1) EMMA obtains and benefits from the generalization capabilities inherent in the SOTA LLM agent through cross-modality imitation learning; (2) Our work sheds novel insights on using LLM feedback to train more versatile and generalizable embodied agents.

\section{Related Work}
\textbf{Agents based on Foundation Models.}
Recent research has increasingly focused on harnessing the capabilities of large pre-trained foundation models to build AI agents~\citep{corr/fudan-agent-survey,yang2023foundation}. These models (e.g., LLMs), benefiting from their commonsense knowledge inherited from Internet-scale pretraining, are able to reason actions according to descriptions of the external environments. For example, given a set of task instructions, LLMs can be elaborately prompted to perform as agents generating high-level step-by-step plans~\cite{icml/HuangAPM22,brohan2023can,arxiv/reflexion,iclr/react,wu2023autogen}, and each step can be parsed into a sequence of robotic actions that are executed via pretrained policies or available APIs~\citep{rss/rt-1,icra/CaP,wang2023voyager}. Furthermore, by using VLMs~\citep{corr/chunyuan_vlm_survey}, plans can also be conditioned on visual inputs that are transformed into language descriptions~\citep{huang2023voxposer,icml/distill_vlm,brohan2023can,gao2023pg-vlm} or token embeddings aligned with LLMs~\citep{icml/palm-e,corr/embodiedgpt,corr/rt-2,yang2023octopus}.
However, existing foundation models are usually pretrained on static text or text-image datasets and thus may struggle to align with the dynamics of the world.
To bridge this gap, we study how to finetune a VLM to be an embodied agent dynamically aligned with the world by distilling the cross-modality knowledge from an LLM expert. The work most closely related to ours is EUREKA~\citep{ma2023eureka}, which also explores using source information provided by the simulator as the context of an LLM to aid agent training. Instead of directly mimicking the output of the LLM as we did, EUREKA harnesses the coding LLM to generate a desired reward function for a given task and optimizes a policy against the reward function using RL, leading to a more complex and expensive training procedure~\citep{icml/trpo,schulman2017ppo,arxiv/dpo}. \looseness -1

\noindent\textbf{Imitation Learning.}
Imitation learning is the study of algorithms that improve performance by mimicking an expert’s decisions and behaviors. We summarize three main categories of existing methods in the following: (1) behavior cloning (BC), (2) inverse reinforcement learning (IRL), and (3) the combination of imitation and reinforcement learning. 
The na\"ive BC~\citep{bain1995bc} ignores the change in distribution and simply trains a policy that only performs well under the distribution of states visited by the expert. Following works, such as dataset aggregation~\citep{aistats/dagger} or policy aggregation~\citep{ml/searn,jmlr/smile}, have been proposed to address the limitations of BC. Another line of work, IRL, is a more complicated algorithm framework that learns the reward function from expert demonstrations and then improves the policy using RL with the learned reward. A representative method in this category is generative adversarial imitation learning (GAIL)~\citep{nips/gail}, in which a policy and a discriminator compete with each other in order to maximize the likelihood of the policy's behavior matching the expert. 
The third category of methods usually leverages an IL policy to initialize RL and continues to boost its performance via online collected data from RL~\citep{icml/lols,ross2014reinforcement,iclr/thor}. This simple combination significantly improves RL's sample efficiency and IL's upper performance bound constrained by the expert. 
Nevertheless, all of the above methods assume that the expert and the imitator understand the world in the same modality, and thus overlook the fact that the complementary knowledge from other modalities often boosts the model's accuracy and generalization dramatically~\citep{icml/clip,iclr/XueGRZ23}.

%\vspace{-10pt}
\section{Conclusion}
%\vspace{-5pt}
We create \textbf{EMMA}, an \textbf{E}mbodied \textbf{M}ulti-\textbf{M}odal \textbf{A}gent, by finetuning a VLM in an embodied visual world with interactive imitation learning from an LLM expert in a parallel text world, who produces better actions and retrospective feedback to VLM's trajectories. Such imitation learning exhibits substantial advantages over vision or VLM policies directly finetuned in the visual world or finetuned by behavior cloning of a rule-based expert, and SOTA API VLMs such as GPT-4V(ision). As a result, EMMA achieves a comparable success rate and much better robustness in the noisy visual world than its LLM teacher in the easier text world. Furthermore, EMMA shows powerful generalization to open-vocabulary and free-form task instructions, highlighting its potential in real-world scenarios.

\section*{Acknowledgment}
This work is partially supported by the National Key R\&D Program of China under the Grant No. (2023YFE0106300) and (2017YFC0804002), Shenzhen Fundamental Research Program under the Grant No. (JCYJ20200109141235597), and Program for Guangdong Introducing Innovative and Entrepreneurial Teams under Grant No. (2017ZT07X386).
% {   \clearpage
{
    \small
    \bibliographystyle{ieeenat_fullname}
    \bibliography{main}
}

\clearpage

\setcounter{page}{1}
\onecolumn
\section*{Appendix}
% \maketitlesupplementary
% \onecolumn

\begin{figure*}[htb]
% \vspace{-0.3em}
    \centering
    \includegraphics[width=0.85\textwidth]{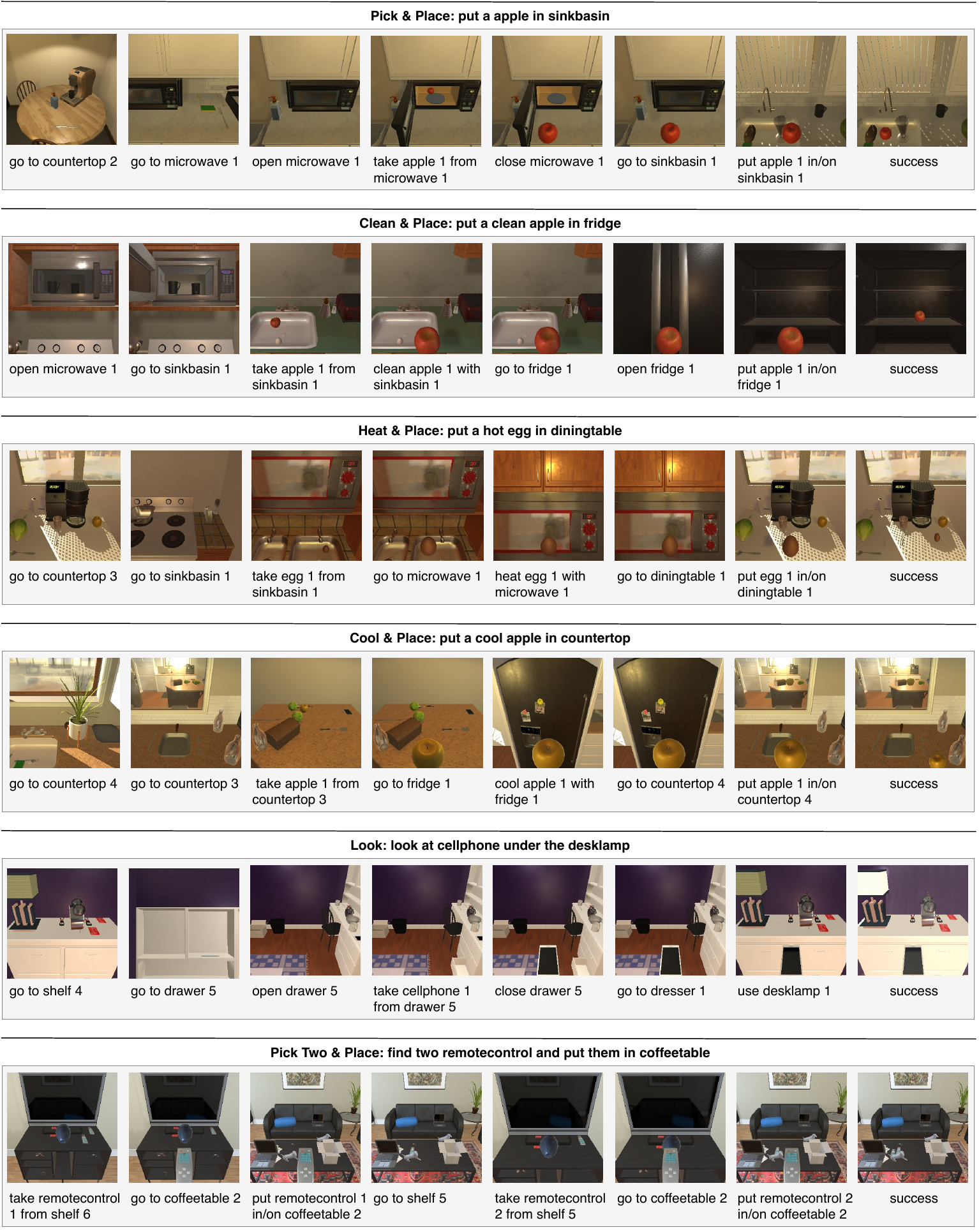}
    % \vspace{-1.5em}
    \caption{\textbf{Visualized task examples of ALFWorld~\citep{iclr/alfworld}.~}This benchmark adopts diverse household scenes developed by the Ai2Thor~\citep{kolve2017ai2thor} environment, in which all objects can be relocated to different positions based on placeable surface areas and class constraints, enabling the procedural generation of a large pool of new tasks via recombining different objects and goal positions.}
    \label{fig:full_example_tasks}
% \vspace{-1.5em}
\end{figure*}

\section{Full Prompts for LLM Expert}
\label{appdix:prompts_llm}
In this section, we provide all LLM prompts for the training procedure (Alg.~\ref{alg:training}) of EMMA. We adopt the prompting technique developed by ReAct~\citep{iclr/react} but ignore the reasoning traces, i.e., ``think'' steps, when executing imitation learning between EMMA and the LLM actor. After each trial $i$, the retrospective feedback $\mathcal{P}_{i}$ generated by the LLM critic will be appended to long-term memory $\mathcal{P}$. In practice, we bound $\mathcal{P}$ by a maximum number of stored feedback (usually set to 1-3) to adhere to the max context length of the LLM.

\begin{table*}[htb]
\scriptsize
\begin{minipage}{0.9\linewidth}
An example of the prompt adopted by LLM actor for generating actions
\centering
\ttfamily
\begin{tabular}{p\linewidth}
\midrule
\textbf{\{environment:} You are in the middle of a room. Looking quickly around you, you see a cabinet 4, a cabinet 3, a cabinet 2, a cabinet 1, a countertop 1, a garbagecan 1, a handtowelholder 2, a handtowelholder 1, a sinkbasin 2, a sinkbasin 1, a toilet 1, a toiletpaperhanger 1, and a towelholder 1. \\
\textbf{task instruction:} Your task is to: put some spraybottle on toilet. \\
> think: To solve the task, I need to find and take a sparybottle, then put it on toilet.\\
> think: First I need to find a spraybottle. A spraybottle is more likely to appear in cabinet (1-4), countertop (1), toilet (1), sinkbasin (1-2), garbagecan (1). I can check one by one, starting with cabinet 1.\\
> go to cabinet 1\\
On the cabinet 1, you see a cloth 1, a soapbar 1, a soapbottle 1.\\
> go to cabinet 2\\
The cabinet 2 is closed.\\
> open cabinet 2\\
You open the cabinet 2. The cabinet 2 is open. In it, you see a candle 1, and a spraybottle 2.\\
> think: Now I find a spraybottle (2). Next, I need to take it.\\
> take spraybottle 2 from cabinet 2\\
You pick up the spraybottle 2 from the cabinet 2.\\
> think: Now I take a spraybottle (2). Next, I need to put it in/on toilet 1.\\
> go to toilet 1\\
On the toilet 1, you see a soapbottle 2.\\
> put spraybottle 2 in/on toilet 1\\
You put the spraybottle 2 in/on the toilet 1.\textbf{\} few-shot examples $\mathbf{\times}$ $\mathbf{n}$}\\
\\
\textbf{long-term memory $\mathcal{P}$:} [...]\\
\textbf{target environment:} ... \\
\textbf{task instruction:} ... \\
> \\
\midrule
\end{tabular}
\end{minipage}%
\label{tab:prompt_actor}
\end{table*}

\begin{table*}[htb]
\scriptsize
\begin{minipage}{0.9\linewidth}
An example of the prompt adopted by LLM critic for generating retrospection
\centering
\ttfamily
\begin{tabular}{p\linewidth}
\midrule
\textbf{\{previous trial:} You are in the middle of a room. Looking quickly around you, you see a cabinet 10, a cabinet 9, a cabinet 8, a cabinet 7, a cabinet 6, a cabinet 5, a cabinet 4, a cabinet 3, a cabinet 2, a cabinet 1, a coffeemachine 1, a countertop 2, a countertop 1, a diningtable 2, a diningtable 1, a drawer 3, a drawer 2, a drawer 1, a fridge 1, a garbagecan 1, a microwave 1, a sinkbasin 1, a stoveburner 4, a stoveburner 3, a stoveburner 2, a stoveburner 1, and a toaster 1.\\
Your task is to: heat some mug and put it in coffeemachine.\\
> go to countertop 1\\
On the countertop 1, you see a butterknife 1, a cup 1, a fork 2, and a mug 1.\\
> take mug 1 from countertop 1\\
You pick up the mug 1 from the countertop 1.\\
> go to stoveburner 1\\
On the stoveburner 1, you see a pan 2.\\
> examine stoveburner 1\\
On the stoveburner 1, you see a pan 2.\\
> examine stoveburner 1\\
On the stoveburner 1, you see a pan 2.\\
> examine stoveburner 1\\
On the stoveburner 1, you see a pan 2.\\
> examine stoveburner 1\\
On the stoveburner 1, you see a pan 2.\\
STATUS: FAIL\\
\textbf{retrospection}: You were stuck in a loop in which you continually examined stoveburner 1 instead of heating mug 1 with stoveburner 1. You should have taken mug 1 from countertop 1, then heated it with stoveburner 1, then put it in coffeemachine 1. It did not help to execute two identical actions in a row. You will try to execute a different action if You am stuck in a loop again.\textbf{\} few-shot examples $\times$ $\mathbf{n}$}\\
\\
\textbf{current trial:} ...\\
\textbf{retrospection:} \\
\midrule
\end{tabular}
\end{minipage}%
\label{tab:prompt_critic}
\end{table*}

\section{Parallel TextWorld}
\label{appdix:tw_details}
While the idea of parallel TextWorld is heavily inspired by previous work~\citep{cvpr/alfred,iclr/alfworld}, we have enhanced the TextWorld engine to create text-based equivalents of each visual environment for training and evaluating language-based agents. This enhancement involves utilizing a combination of the PDDL~\citep{1998pddl} and Fast Downward~\citep{helmert2006fast} to maintain and update the textual state of the simulated environments. Based on the metadata provided by the simulator, we represent a visual state as a list of attributes. These attributes detail the relationships among various entities in the environment, such as positions, goals, and objects. Note that all these attributes, variables, and rules are defined within the framework of PDDL.
% The idea of parallel TextWorld is heavily inspired by previous work~\citep{cvpr/alfred,iclr/alfworld}. In order to procedurally generate textual environments for training and evaluating language-based agents, we extend TextWorld engine for creating text-based equivalents of each visual environment. Specifically, we use the combination of PDDL~\citep{1998pddl} and Fast Downward~\citep{helmert2006fast} to maintain and update the current state of the simulated environments. Based on the metadata from simulator, a textual state can be represented by a list of attributes which defines the relations between the entities, e.g., positions, goals, and objects, present in the simulated environment. Note that all attributes, variables, and rules are defined using the PDDL language.

\section{Training Details}
\label{appdix:training_details}
We provide hyperparameters used for training EMMA in Table~\ref{tab:hyperparameters}. These hyperparameters are largely derived from those proposed for finetuning InstructBLIP model~\citep{corr/instructblip}. When training, we only update the parameters of linear projection layer while keeping other components frozen, as done during instruction tuning for many existing work~\citep{corr/minigpt4,corr/llama-adaptor}. We use the AdamW optimizer~\citep{loshchilov2017decoupled} with a linear warmup of the learning rate, followed by a linear decay with a minimum learning rate of 0. Moreover, we remove the instruction input of Q-Former, which is used in InstructBLIP, and find this improves performance cross all experiments. Our implementation is heavily inspired by the LAVIS library~\citep{li-etal-2023-lavis} so the training and evaluation processes use the standard procedure provided by LAVIS.

\begin{table}[htb]
% \vspace{-0.3cm}
% \vspace{-0.3cm}
% \vskip 0.15in
\caption{Hyperparameters of EMMA for ALFWorld experiments}
\label{tab:hyperparameters}
\begin{center}
\begin{small}
\begin{tabular}{lcc}
\toprule
\textbf{Hyperparameter} & & \textbf{Value}  \\
\midrule
\multicolumn{3}{c}{\textbf{EMMA's Architecture}} \\
LLM decoder       && Vicuna-7b-v1.1~\citep{zheng2023vicuna} \\
Image encoder     && ViT-L~\citep{icml/clip} \\
Q-Former          && $\text{BERT}_{\text{base}}$~\citep{naacl/bert} \\
Pretrained weights && InstructBLIP~\citep{corr/instructblip} \\
Number of query tokens && $32$ \\
Q-Former text input && False \\
Max text length   && 1024 \\
Image resolution  && 224 \\
\midrule
\multicolumn{3}{c}{\textbf{Behavior Cloning}} \\
Finetuning epochs && $6$ \\
Warmup steps      && $1000$ \\
Learning rate     && $10^{-5}$ \\
Batch size        && $128$ \\
AdamW $\beta$     && $(0.9,0.999)$ \\
Weight decay      && $0.05$ \\
Drop path         && $0$ \\
Inference beam size && $5$ \\
\midrule
\multicolumn{3}{c}{\textbf{Imitation Learning}} \\
Base model for LLM expert     && text-davinci-003 \\
Prompts for LLM expert && refer to Sec.~\ref{appdix:prompts_llm} \\
Number of trials       && $12$ \\
Episode length     && $30$ \\
Size of long-term memory        && $3$ \\
Learning rate      && $5\times 10^{-6}$ \\
Warmup steps      && $300$ \\
Batch size         && $16$ \\
Training epochs per trial && $5$ \\
DPO $\beta$       && $0.1$ \\
\bottomrule
\end{tabular}
\end{small}
\end{center}
% \vskip -0.1in
\end{table}

\begin{figure*}[htb!]
% \vspace{-0.3em}
    \centering
    \includegraphics[width=0.9\textwidth]{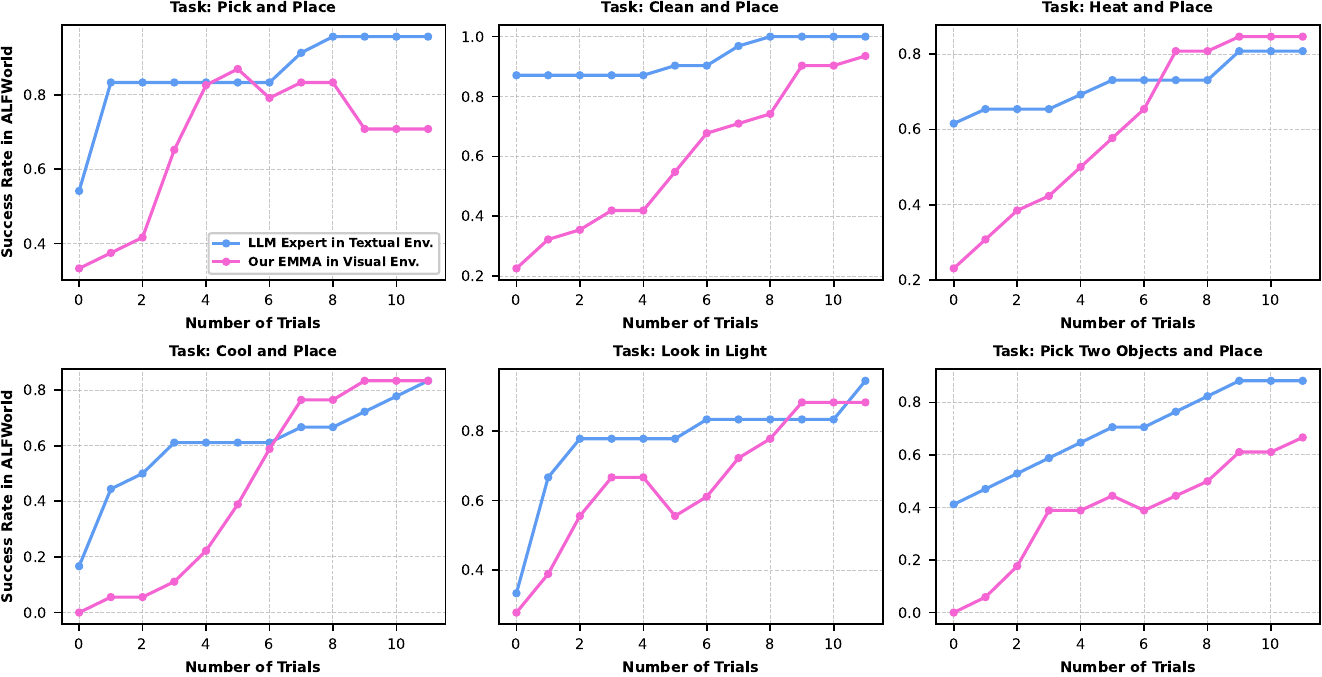}
    % \vspace{-1.5em}
    \caption{\looseness=-1 \small \textbf{Comparison of success rate between EMMA and the LLM expert.} As the number of trials increases, the gap between the two agents decreases, and EMMA even outperforms or matches the expert in some tasks (e.g., ``Heat and Place'' and ``Cool and Place''), indicating the effectiveness of cross-modality imitation learning.}
    \label{fig:imitate_curve_sep}
% \vspace{-1.5em}
\end{figure*}

\begin{figure*}[htb!]
% \vspace{-0.3em}
    \centering
    \includegraphics[width=0.9\textwidth]{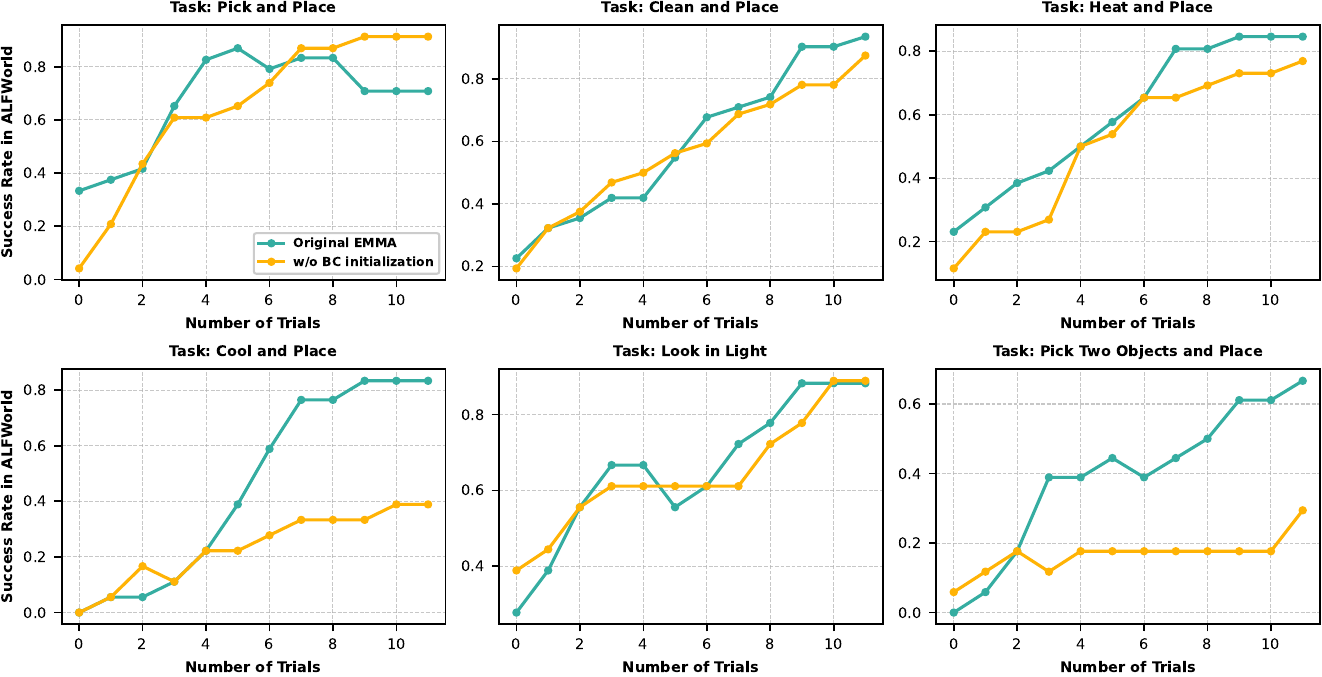}
    % \vspace{-1.5em}
    \caption{\looseness=-1 \small \textbf{Ablation study.} The performance of ``EMMA w/o BC initialization'' is consistently worse than the original EMMA.}
    \label{fig:ablation_bc_separate}
% \vspace{-1.5em}
\end{figure*}

\begin{figure*}[htb!]
% \vspace{-0.3em}
    \centering
    \includegraphics[width=\textwidth]{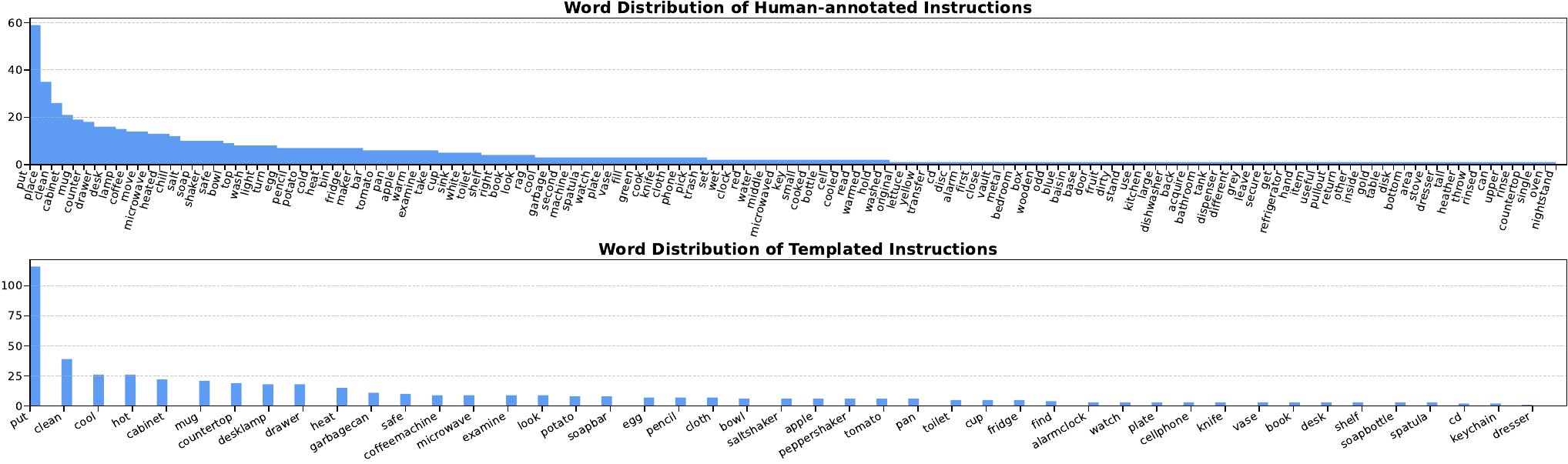}
    % \vspace{-1.5em}
    \caption{\looseness=-1 \small \textbf{Vocabulary Distributions.} Frequency distribution of all words for human-annotated and templated task instructions. The diversity of human-annotated instructions presents a significant challenge for the generalization abilities of agents.}
    \label{fig:all_words_distribution}
% \vspace{-1.5em}
\end{figure*}
\clearpage

\section{Collection of Demonstration Dataset}
\label{appdix:dataset}
Fine-tuning pretrained VLMs on a pre-collected demonstration dataset via behavior cloning is a critical step, enabling these models to comprehend and follow the unique grammar of ALFWorld as well as to develop a basic ``gamesense''. 
However, the number of task instructions in the original ALFWorld~\citep{cvpr/alfred} is too limited to yield sufficient data for fine-tuning these large pretrained VLMs effectively. 
Hence, we propose an automated pipeline, which leverages \textit{text-davinci-003} and a rule-based planner to generate a large amount of new instructions and their resulting expert demonstrations, respectively.

To generate a diverse set of new task instructions, we harness the in-context learning capabilities of LLM. Our procedure begins with extracting detailed descriptions from the ALFWorld benchmark for each environment, providing comprehensive information on the number and functional attributes of all items. Then, based on the types of room in these environments, we design different prompts that aim at inducing the LLM to generate task instructions aligned with the features of the target environment. An example of these prompts is detailed in Table~\ref{tab:prompt_kitchen}. 
For each generated task instruction, we gather demonstrations $\{x_{\text{task}},s^{t}_{v},x^{t}_{a}\}_{t=0}^{T}$ using a rule-based planner devised by ALFWorld. It's important to note that this planner operates with an unfair advantage: it considers the environment as fully observable and has complete information of world dynamics, relying on metadata that is not accessible to the agent during training. In summary, our dataset comprises 15247 expert demonstration episodes, amounting to 178585 image-text pairs.

\begin{table*}[htb]
\caption{An example of the prompt for generating new task instructions in the kitchen}
\label{tab:prompt_kitchen}
\scriptsize
\centering
\resizebox{0.85\linewidth}{!}{
\ttfamily
\begin{tabular}{p\linewidth}
\midrule
Q:\\
\textbf{environment:} You are in the middle of a room. Looking quickly around you, you see a cabinet, a countertop, a cabinet, a countertop, a drawer, a drawer, a drawer, a stoveburner, a stoveburner, a drawer, a stoveburner, a stoveburner, a cabinet, a cabinet, a microwave, a cabinet, a cabinet, a cabinet, a sink, a sinkbasin, a fridge, a toaster, a coffeemachine, a cabinet, a drawer, a drawer, a drawer, a drawer, a shelf, a shelf, a countertop, a shelf, a drawer, and a garbagecan.\\
\textbf{object dictionary:} all of operable objects are listed in the following dictionary with a consistent format \{type of operation: \{object name: number of objects\}\}: \{'pickupable': \{'dishsponge': 3, 'apple': 2, 'butterknife': 3, 'peppershaker': 2, 'saltshaker': 3, 'bowl': 2, 'spatula': 2, 'pot': 3, 'winebottle': 3, 'statue': 2, 'creditcard': 3, 'plate': 2, 'pan': 2, 'kettle': 3, 'soapbottle': 3, 'potato': 3, 'fork': 2, 'bread': 2, 'knife': 3, 'glassbottle': 3, 'book': 1, 'tomato': 1, 'vase': 2, 'egg': 1, 'papertowelroll': 1, 'cup': 1, 'lettuce': 1, 'spoon': 1, 'mug': 1\}, 'sliceable': \{'apple': 2, 'potato': 3, 'bread': 2, 'tomato': 1, 'egg': 1, 'lettuce': 1\}, 'receptacle': \{'bowl': 2, 'pot': 3, 'plate': 2, 'pan': 2, 'stoveburner': 4, 'drawer': 9, 'countertop': 3, 'cabinet': 9, 'microwave': 1, 'shelf': 3, 'toaster': 1, 'garbagecan': 1, 'cup': 1, 'fridge': 1, 'coffeemachine': 1, 'sinkbasin': 1, 'mug': 1\}, 'canFillWithLiquid': \{'bowl': 2, 'pot': 3, 'winebottle': 3, 'kettle': 3, 'glassbottle': 3, 'houseplant': 1, 'cup': 1, 'mug': 1\}, 'breakable': \{'bowl': 2, 'winebottle': 3, 'statue': 2, 'plate': 2, 'glassbottle': 3, 'window': 1, 'vase': 2, 'egg': 1, 'cup': 1, 'mug': 1\}, 'dirtyable': \{'bowl': 2, 'pot': 3, 'plate': 2, 'pan': 2, 'cup': 1, 'mug': 1\}, 'openable': \{'kettle': 3, 'drawer': 9, 'cabinet': 9, 'book': 1, 'microwave': 1, 'fridge': 1\}, 'cookable': \{'potato': 3\}, 'toggleable': \{'stoveknob': 4, 'microwave': 1, 'toaster': 1, 'coffeemachine': 1, 'lightswitch': 1, 'faucet': 1\}\}\\
\textbf{generate 30 diverse tasks based on the environment description and object dictionary.}\\
A:\\
Task 1: pick\_clean\_then\_place\_in\_recep, put a clean pan in fridge\\
Task 2: pick\_and\_place\_simple, put a apple in countertop\\
Task 3: pick\_two\_obj\_and\_place, put two apple in garbagecan\\
Task 4: pick\_heat\_then\_place\_in\_recep, put a hot apple on plate\\
Task 5: pick\_cool\_then\_place\_in\_recep, put a cool apple in countertop\\
Task 6: pick\_and\_place\_simple, put a mug in coffeemachine\\
Task 7: pick\_two\_obj\_and\_place, put two creditcard in plate\\
Task 8: pick\_clean\_then\_place\_in\_recep, put a clean mug in coffeemachine\\
Task 9: pick\_heat\_then\_place\_in\_recep, put a hot mug on plate\\
Task 10: pick\_cool\_then\_place\_in\_recep, put a cool mug in plate\\
Task 11: pick\_and\_place\_simple, put a peppershaker in cabinet\\
Task 12: pick\_two\_obj\_and\_place, put two peppershaker in shelf\\
Task 13: pick\_clean\_then\_place\_in\_recep, put a clean fork in pot\\
Task 14: pick\_heat\_then\_place\_in\_recep, put a hot fork on bowl\\
Task 15: pick\_cool\_then\_place\_in\_recep, put a cool fork in plate\\
Task 16: pick\_and\_place\_simple, put a statue in countertop\\
Task 17: pick\_two\_obj\_and\_place, put two bowl in microwave\\
Task 18: pick\_clean\_then\_place\_in\_recep, put a clean potato in fridge\\
Task 19: pick\_heat\_then\_place\_in\_recep, put a hot potato on plate\\
Task 20: pick\_cool\_then\_place\_in\_recep, put a cool potato in pot\\
Task 21: pick\_and\_place\_simple, put a egg in countertop\\
Task 22: pick\_two\_obj\_and\_place, put two bread in microwave\\
Task 23: pick\_clean\_then\_place\_in\_recep, put a clean egg in garbagecan\\
Task 24: pick\_heat\_then\_place\_in\_recep, put a hot egg on bowl\\
Task 25: pick\_cool\_then\_place\_in\_recep, put a cool egg in pan\\
Task 26: pick\_and\_place\_simple, put a pan in stoveburner\\
Task 27: pick\_two\_obj\_and\_place, put two pot in stoveburner\\
Task 28: pick\_clean\_then\_place\_in\_recep, put a clean tomato in coffeemachine\\
Task 29: pick\_heat\_then\_place\_in\_recep, put a hot tomato on plate\\
Task 30: pick\_cool\_then\_place\_in\_recep, put a cool tomato in plate\\
\\
Q:\\
\textbf{environment:} ...\\
\textbf{object dictionary:} ...\\
generate 30 diverse tasks based on the environment description and object dictionary.\\
A:\\
\textbf{...LLM-generated task instructions...}\\
\midrule \\
\end{tabular}
}%
\end{table*}

\end{document}